\documentclass[sigconf]{acmart}
\pdfoutput=1
\usepackage{booktabs} % For formal tables
\usepackage{subfigure}%in preamble
\usepackage{multirow}
\usepackage{graphicx}
\usepackage{algorithm}    
\usepackage{algorithmic} 
\usepackage{mathrsfs}
\usepackage{enumitem}
\usepackage{balance}
\newcommand{\tabincell}[2]{\begin{tabular}{@{}#1@{}}#2\end{tabular}}

\begin{document}
\title{Supervised Reinforcement Learning with Recurrent Neural Network for Dynamic Treatment Recommendation}
\author{Lu Wang}
\affiliation{%
	\institution{School of Computer Science and Software Engineering\\East China Normal University}
}
\email{joywanglulu@163.com}

\author{Wei Zhang*}\thanks{*Co-corresponding authors.}
\orcid{0000-0001-6763-8146}
\affiliation{%
	\institution{School of Computer Science and Software Engineering\\East China Normal University}
}
\email{zhangwei.thu2011@gmail.com}

\author{Xiaofeng He*} 
\affiliation{%
	\institution{School of Computer Science and Software Engineering\\East China Normal University}
}
\email{xfhe@sei.ecnu.edu.cn}

\author{Hongyuan Zha}
\affiliation{%
	\institution{Georgia Tech}
	%\institution{East China Normal University}
}
\email{zha@cc.gatech.edu}

% The default list of authors is too long for headers.
\renewcommand{\shortauthors}{Lu Wang et al.}
\renewcommand{\shorttitle}{SRL-RNN for Dynamic Treatment Recommendation}

\begin{abstract}
	Dynamic treatment recommendation systems based on large-scale electronic health records (EHRs) become a key to successfully improve practical clinical outcomes. Prior relevant studies recommend treatments either use supervised learning (e.g. matching the indicator signal which denotes doctor prescriptions), or reinforcement learning (e.g. maximizing evaluation signal which indicates cumulative reward from survival rates). However, none of these studies have considered to combine the benefits of supervised learning and reinforcement learning. In this paper, we propose Supervised Reinforcement Learning with Recurrent Neural Network (SRL-RNN), which fuses them into a synergistic learning framework. Specifically, SRL-RNN applies an off-policy actor-critic framework to handle complex relations among multiple medications, diseases and individual characteristics. The ``actor'' in the framework is adjusted by both the indicator signal and evaluation signal to ensure effective prescription and low mortality. RNN is further utilized to solve the Partially-Observed Markov Decision Process (POMDP) problem due to the lack of fully observed states in real world applications. Experiments on the publicly real-world dataset, i.e., MIMIC-3, illustrate that our model can reduce the estimated mortality, while providing promising accuracy in matching doctors' prescriptions. 

\end{abstract}
%\begin{abstract}
%Comorbidity is a pervasive phenomenon and challenge in healthcare, due to the longer and frequent hospitalisations cuased by complex interrelation of diseases, medications and individual characteristics. Prior works either use error signal to match the behaviours of doctors using supervised learning, or use evaluation signal to adapt a treatment plan of one disese as a dynamic treatment (DTR) to recommend time-varying medications and explore optimal treatment using reinforcement learning. We also adapt treatment plan of comorbidity as DTR. Instead, we propose supervised reinforcement learning with recurrent neural network (SRL-RNN), a method that uses error signal and evaluation signal at the same time. In SRL-RNN, an off-policy actor-critic framework is intrgreted to handle large discrete medications, a supervisor of doctor behavior directly adjust the ``actor'' to promise safe actions and a RNN extends SRL to solve POMDP problems. Sufficient experiments on real world EHR datasets demonstrates our method can reduce the mortality in the hospital by up to 4.3\% from a baseline 21\%.
%\end{abstract}

%
% The code below should be generated by the tool at
% http://dl.acm.org/ccs.cfm
% Please copy and paste the code instead of the example below.
%


\maketitle

% end the environment with {table*}, NOTE not {table}!
\section{Introduction}
Treatment recommendation has been studied for a long history.
Specially, medication recommendation systems have been verified to support doctors in making better clinical decisions.
Early treatment recommendation systems match diseases with medications via classification based on expert systems~\cite{Zhuo2016A,Gunlicksstoessel2017A,Almirall2012Designing}. But it heavily relies on knowledge from doctors,
and is difficult to achieve personalized medicine. With the availability of electronic health records (EHRs) in recent years,
there are enormous interests to exploit personalized healthcare data to optimize clinical decision making.
Thus the research on treatment recommendation shifts from knowledge-driven into data-driven. 

The data-driven research on treatment recommendation involves two main branches:
supervised learning (SL) and reinforcement learning (RL) for prescription.
SL based prescription tries to minimize the difference between the recommended prescriptions and indicator signal which denotes doctor prescriptions.
Several pattern-based methods generate recommendations by utilizing the similarity of patients~\cite{zhang2014towards,sun2016data,hu2016data}, 
but they are challenging to directly learn the relation between patients and medications.
Recently, some deep models achieve significant improvements
by learning a nonlinear mapping from multiple diseases to multiple drug categories~
\cite{bajor2017predicting, Zhang2017LEAP, wang2018personalized}.
Unfortunately, a key concern for these SL based models still remains unresolved, i.e, the ground truth of ``good'' treatment strategy
being unclear in the medical literature~\cite{Marik2015The}. 
More importantly, the original goal of clinical decision also considers the outcome of patients
instead of only matching the indicator signal.

\begin{figure}
	%\vspace{-0.8em}
	\includegraphics[width=3.7in]{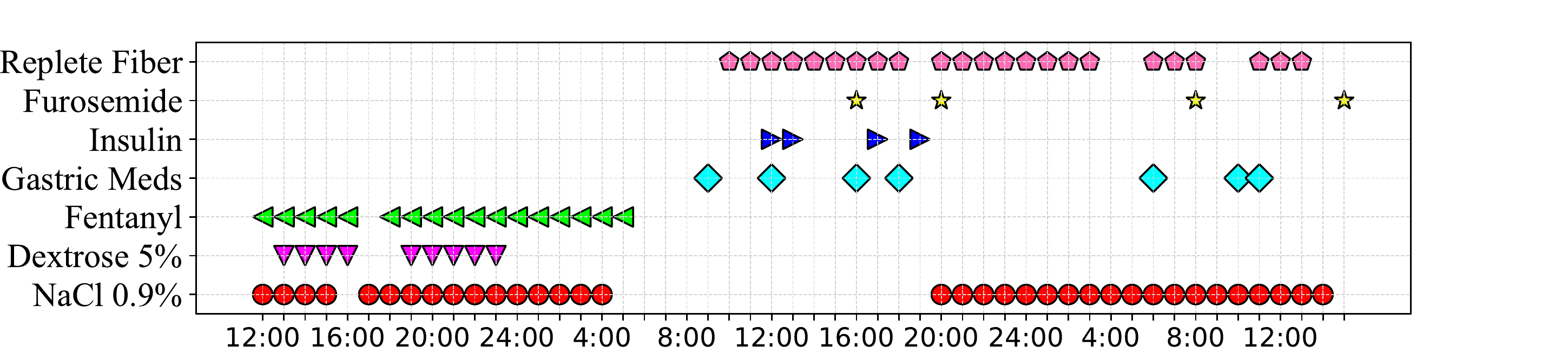}
	%\vspace{-2.5em}
	\caption{The record in MIMIC-3 shows a treatment process of a patient with acute respiratory failure, chronic kidney disease, etc,
		where seven medications are randomly selected.
		The x-axis represents the time-line from admission time 12:00 to the discharge time 17:00 of the third day.
		Each icon represents a prescribed medication for the patient.}
	\label{fig:intro}
	%	\vspace{-1.5em}
\end{figure}

The above issues can be addressed by reinforcement learning for dynamic treatment regime (DTR)~\cite{robins1986new,murphy2003optimal}.
DTR is a sequence of tailored treatments according to the dynamic states of patients,
which conforms to the clinical practice. As a real example shown in Figure \ref{fig:intro},
treatments for the patient vary dynamically over time with the accruing observations.
The optimal DTR is determined by maximizing the evaluation signal which indicates the long-term outcome of patients,
due to the delayed effect of the current treatment and the influence of future treatment choices~\cite{chakraborty2013statistical}. 
With the desired properties of dealing with delayed reward and inferring optimal policy based on non-optimal prescription behaviors, 
a set of reinforcement learning methods have been adapted to generate optimal DTR for life-threatening diseases,
such as schizophrenia, non-small cell lung cancer, and sepsis~\cite{Shortreed2012Estimating,Zhao2011Reinforcement,Nemati2016Optimal}.
Recently, some studies employ deep RL to solve the DTR problem based on large scale EHRs~\cite{weng2017representation,Prasad2017A,raghu2017deep}. 
Nevertheless, these methods may recommend treatments that are obviously different from doctors' prescriptions
due to the lack of the supervision from doctors, which may cause high risk \cite{mihatsch2002risk} in clinical practice.
In addition, the existing methods are challenging for analyzing multiple diseases and the complex medication space. 
%This might be the main reason that most RL based methods are only applied to the simulated or low-risk environments.

In fact, the evaluation signal and indicator signal paly complementary roles~\cite{Barto2002Reinforcement,Clouse1992A}, 
where the indicator signal gives a basic effectiveness and the evaluation signal helps optimize policy.
Imitation learning~\cite{abbeel2004apprenticeship,ratliff2006maximum,ziebart2008maximum,levine2011nonlinear,finn2016guided}
utilizes the indicator signal to estimate a reward function for training robots by supposing the indicator signal is optimal,
which is not in line with the clinical reality. Supervised actor-critic~\cite{Clouse1992A,benbrahim1997biped,barto2004j}
uses the indicator signal to pre-train a ``guardian'' and then combines ``actor'' output and ``guardian'' output
to send low-risk actions for robots. However, the two types of signals are trained separately
and cannot learn from each other.
Inspired by these studies, we propose a novel deep architecture to generate recommendations for more general DTR
involving multiple diseases and medications, called Supervised Reinforcement Learning with Recurrent Neural Network (SRL-RNN).
The main novelty of SRL-RNN is to combine the evaluation signal and indicator signal at the same time to learn an integrated policy. 
More specifically, SRL-RNN consists of an off-policy actor-critic framework to learn complex relations among medications, diseases, and individual characteristics. 
The ``actor'' in the framework is not only influenced by the evaluation signal like traditional RL
but also adjusted by the doctors' behaviors to ensure safe actions. 
RNN is further adopted to capture the dependence of the longitudinal and temporal records of patients for the POMDP problem. 
Note that treatment and prescription are used interchangeably in this paper.

Our contributions can be summarized as follows:
\begin{itemize}
	%\vspace{-0.5em}
	\item We propose a new deep architecture SRL-RNN for handling a more general DTR setting involving multiple diseases and medications.
	It learns the prescription policy by combing both the indicator signal and evaluation signal to avoid unacceptable risks
	and infer the optimal dynamic treatment.
	\item
   SRL-RNN applies an off-policy actor-critic framework to handle complex relations among multiple medications, diseases, and individual characteristics.
   The ``actor'' is adjusted by both the indicator signal and evaluation signal and RNN is further utilized to solve POMDP
   (see Section~\ref{srl-recurrent}).
%	SRL-RNN incorporates an off-policy actor-critic architecture to jointly learn the SL and RL tasks,
%	and further adopt long short term memory (LSTM) to solve the POMDP problem. 
	\item
	Quantitative experiments and qualitative case studies on MIMIC-3 demonstrate that our method can not only reduce
	the estimated mortality in the hospital (see Section~\ref{sec:evaluation-metric})
	by 4.4\%, but also provide better medication recommendation.% (see Section~\ref{reasult}).
%	(see Table~\ref{tab:all})
	%\vspace{-5pt}
\end{itemize}
The rest of this paper is organized as follows. We summarize the related work in Section~\ref{rel}
and provide necessary background knowledge in Section~\ref{bac} for later introduction.
In what follows, our model is specified in Section~\ref{srl}. Experimental results are presented in Section~\ref{exp}.
Finally, we conclude our paper in Section~\ref{con}.

\section{Related  Work}\label{rel}
Early treatment recommendation systems heuristically map diseases into medications based on
expert systems~\cite{Zhuo2016A,Gunlicksstoessel2017A,Almirall2012Designing}. Due to the difficulty of knowledge acquisition,
it comes into data-driven approaches with two branches: supervised learning and reinforcement learning.
In this section, we overview the related studies on data-driven treatment recommendation,
and the methodologies of combing supervised learning and reinforcement learning. 

\textbf{Supervised learning for prescription} focuses on minimizing the difference between
recommended prescriptions and doctors' prescriptions. 
Both Cheerla et al.~\cite{Cheerla2017MicroRNA}
and Rosen et al.~\cite{Rosen2008Selecting}
proposed to utilize genomic information for recommending
suitable treatments for patients with different diseases.
However, genomic information is not widely spread and easy to acquire.
In order to leverage massive EHRs to improve treatment recommendation,
several pattern-based methods generate treatments by the similarity among patients~\cite{zhang2014towards,sun2016data,hu2016data}.
Nevertheless, these methods are challenging to directly learn the relationship between patients and medications. 
Furthermore, it is challenging to calculate the similarities between patients' complex longitudinal records.
Recently, two deep models are proposed to learn a nonlinear mapping from diseases to drug categories based on EHRs,
and achieve significant improvements. Bajor et al.~\cite{bajor2017predicting} adopted a GRU model to predict
the total medication categories given historical diagnosis records of patients.
Zhang et al.~\cite{Zhang2017LEAP} proposed a deep model to not only learn the relations between multiple diseases
and multiple medication categories, but also capture the dependence among medication categories.
In addition, Wang et al.~\cite{wang2018personalized} utilized a trilinear model to integrate
multi-source patient-specific information for personalized medicine.

A major concern for these SL-based prescription is that the behaviors of doctors are prone to be imperfect.
Due to the knowledge gap and limited experiences of doctors, the ground truth of ``good'' treatment strategy
is unclear in the medical literature~\cite{Marik2015The}.
To address this issue, we prefer to use RL that is well-suited to infer optimal policies based on non-optimal
prescriptions.

\textbf{Reinforcement learning for prescription} gives the treatments by maximizing the cumulated reward, 
where the reward can be assessment scores of disease or survival rates of patients. 
Daniel et al.~\cite{Susan2011Informing} employed tabular Q-learning to recommend medications for schizophrenia patients on real clinical data.
Zhao et al.~\cite{Zhao2011Reinforcement} applied fitted Q-learning to discover optimal individualized medications for non-small cell lung cancer (NSCLC) based on simulation data, where a support vector regression (SVR) model is used to estimate Q-function.
Nemati et al.~\cite{Nemati2016Optimal} leveraged a combination of Hidden Markov Models and deep Q-networks to predict optimal
heparin dosing for patients in ICU under a POMDP environment.
Applying these approaches to clinical practice is challenging for their use of relatively small amount of data.
Most recently, based on large scale available EHRs, Weng et al.~\cite{weng2017representation} combined sparse autoencoder 
and policy iteration to predict personalized optimal glycemic trajectories for severely ill septic patients, which reduces 6.3\% mortality.
Prasad et al.~\cite{Prasad2017A} used Q-learning to predict personalized sedation dosage and ventilator support.
Raghu et al.~\cite{raghu2017deep} employed dual double deep Q-learning with continuous-state spaces to recommend optimal
dosages of intravenous fluid and maximum vasopressor for sepsis. 

However, without knowledgeable supervisors, the system may recommend treatments that are significantly different from doctors', which may cause unacceptable risks~\cite{mihatsch2002risk}.
Besides, these value based methods are hard to handle multiple diseases and complex medication space.
%This might be the main reason that most of RL methods are only applied into simulated or low-risk enviroments.

\textbf{Methods of combing SL and RL} utilize expert behaviors to accelerate reinforcement learning
and avoid the risks of actions. Common examples are 
imitation learning~\cite{abbeel2004apprenticeship,ratliff2006maximum,ziebart2008maximum,levine2011nonlinear,finn2016guided}
and supervised actor-critic~\cite{Clouse1992A,benbrahim1997biped,barto2004j}.
Given the samples of expert trajectories, imitation learning requires th skill of estimating a reward function
in which the reward of expert trajectories should enjoy the highest rewards.
Recently, imitation learning combines deep learning to produce successful applications in robotics~\cite{finn2016guided,levine2016end}.
However, these methods assume expert behaviors are optimal, which is not in line with clinical reality.
Besides, most of them learn the policy based on the estimated reward function, instead of directly telling learners how to act.
Supervised actor-critic uses the indicator signal to pre-train a ``guardian'', and sends a low-risk action for robots by
the weighted sum of ``actor'' output and ``guardian'' output, but each signal cannot learn from each other in the training process.
Since this type of model requires much more expert behaviors, it has limited applications.
In this paper, we focus on combining RL and SL by utilizing the large amount of doctors' prescription behaviors. 
Our proposed model SRL-RNN is novel in that: (1) we train ``actor'' with the indicator signal and evaluation signal jointly
instead of learning ``guardian'' and ``actor'' separately; (2) SRL-RNN combines an off-policy RL and classification based SL models, 
while supervised actor-critic combines an on-policy RL and regression based SL models; and (3) SRL-RNN enables capturing the dependence of
longitudinal and temporal records of patients to solve the problem of POMDP.

\section{Background\label{bac}}

In this section, we give a definition of the Dynamic Treatment Regime (DTR) problem
and an overview of preliminaries for our model.
Some important notations mentioned in this paper are summarized in Table~\ref{tab:notation}.
%The core of our approach is a combination of reinforcement learning task deep deterministic policy gradient (DDPG)\cite{lillicrap2015continuous} and supervised learning task (multi-lable classification), each of which achieves better performance than applying only one of them. 

\subsection{Problem Formulation}
In this paper, DTR is modeled as a \emph{Markov decision process} (MDP) with finite time steps and a deterministic policy
consisting of an action space $\mathcal{A}$, a state space $\mathcal{S}$, and a reward function $r:\mathcal{S}\times \mathcal{A} \rightarrow \mathbb{R}$.
At each time step $t$, a doctor observes the current state $s_t\in \mathcal{S}$ of a patient,
chooses the medication $\hat{a}_t\in \{0,1\}^K$ from candidate set $\mathcal{A}$ based on an unknown policy $\hat{\mu}(s)$, and receives a reward $r_t$.
Given the current observations of diseases, demographics, lab values, vital signs, and the output event of the patient which indicate the state $s_t$ , our goal is to learn a policy $\mu_ \theta(s)$ to select an action (medication) $a_t$ by maximizing the sum of discounted rewards (return)
from time step $t$, which is defined as  $R_t=\sum_{i=t}^{T}\gamma^{(i-t)}r(s_i,a_i)$, and simultaneously minimizes the difference from
clinician decisions $\hat{a}_t$. 

There are two types of methods to learn the policy: the value-based RL learning a greedy policy $\mu(s)$ and the policy gradient RL 
maintaining a parameterized stochastic policy $\pi_\theta(a\vert s)$ or a deterministic policy $\mu_\theta(s)$ (see~\ref{sec2} for details),
where $\theta$ is the parameter of the policy.

\begin{figure*}
	\includegraphics[width=6in]{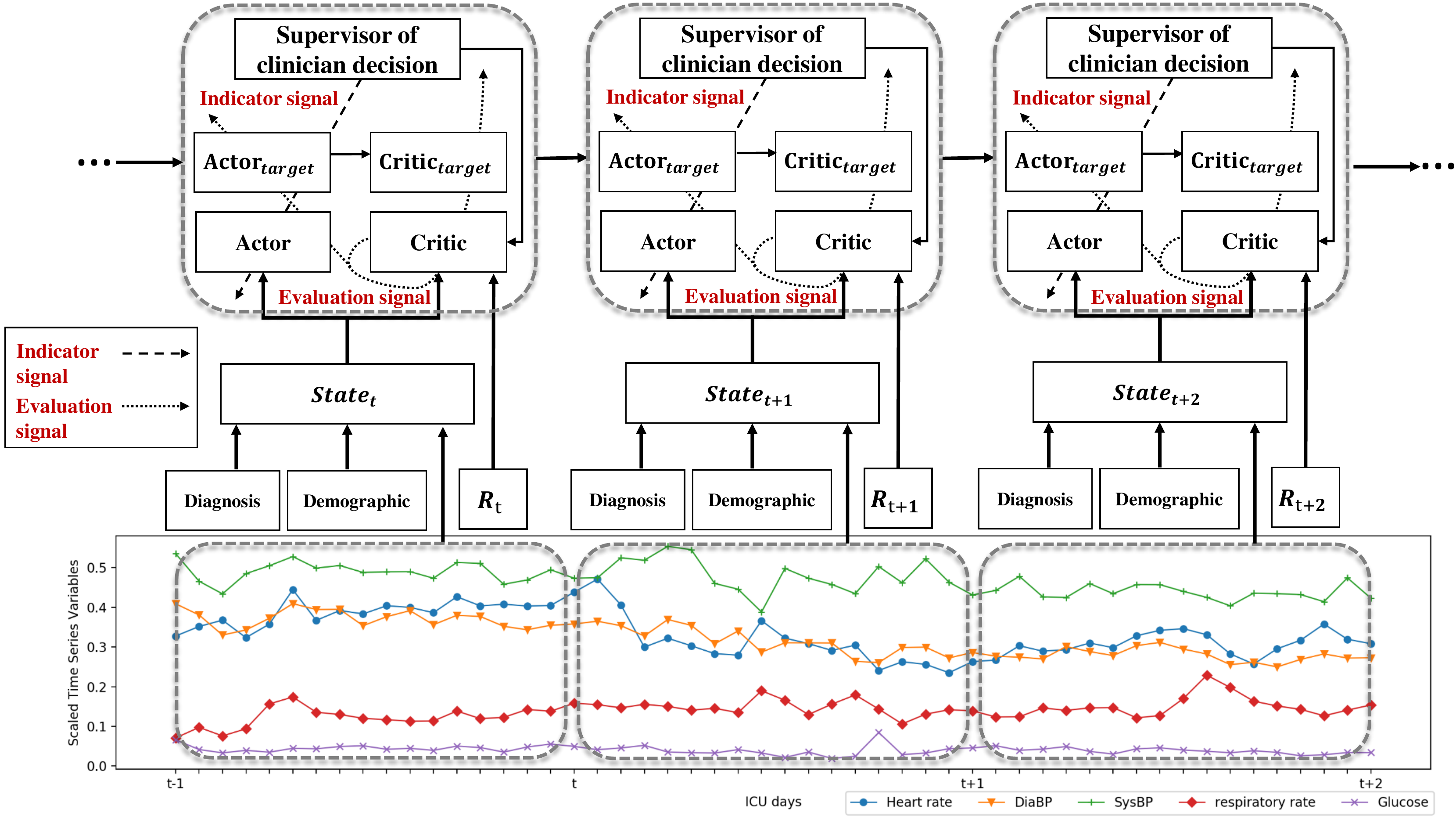}
	\vspace{-10pt}
	\caption{The general framework of supervised reinforcement learning with recurrent neural network. Solid arrows indicate the input diagnoses,
		demographics, rewards, clinician decisions timeseries variables, and historical states.
		Dashed arrows represent the indicator signal and evaluation signal to update the actor ($actor_{target}$) network and critic ($critic_{target}$) network. }
%	\vspace{-1em}
	\label{fig:model}
\end{figure*}

\begin{table}[!t]
	{\centering}
	\small 
	\caption{Important notations.} \label{tab:notation}
%	\vspace{-1.5em}
	\begin{tabular}{|c|c|}
		\hline
		\textbf{Notation}&	\textbf{Description}\\\hline
		$K$ &the number of medications or medication categories\\\hline
		$\mathcal{A},a$&\tabincell{c}{action space $\mathcal{A}\in \{0,1\}^K$, $a\in\mathcal{A} ,a_k\in  \{0,1\} $ indicates \\whether the $k$-th medication is chosen for a patient}\\\hline
		$\mathcal{S},s$ &\tabincell{c}{state space $\mathcal{S}$, $s\in\mathcal{S}$ consists of static \\demographic information, lab values, vital signs, etc. }\\\hline
		$r$&\tabincell{c}{reward function, if a patient survives, r=15, \\ if the patient dies, r=-15 , and r=0 otherwise }\\\hline
		$\gamma$ &\tabincell{c}{$\gamma\in[0,1]$ is a discount factor to balance the \\importance of immediate and future rewards}\\\hline
		%& $E$& enviroment \\\hline
		$\mu_ \theta(s)$ &deterministic policy learned from policy gradient\\\hline
		$\pi_\theta(a\vert s)$ & stochastic policy learned from policy gradient\\\hline
		$\mu(s)$ & a greedy policy learned from Q-learning\\\hline
		$\hat{\mu}(s)$& unknown policy of a doctor\\\hline
		$\beta(a\vert s)$ & \tabincell{c}{behavior policy,\\ generate trajectories for off-policy learning}\\\hline
		%	& $p_1(s_1)$&initial state distribution\\\hline
		%&  $p(s'\vert s)$&\tabincell{c}{$p(s\rightarrow s',t,\pi)$,the probability at state $s'$ \\after transition of t time steps from $s$}\\\hline
		%& $\rho^\pi(s')$&\tabincell{c}{the discounted state distribution \\$\rho^\pi(s'):=\int_\mathcal{S}\sum^T_{t=1}\gamma^{t-1}p_1(s)p(s\rightarrow s',t,\pi)\texttt{d}s$}\\\hline
		\tabincell{c} {$Q^\mu(s,a)$,\\$Q^\pi(s,a)$}& \tabincell{c}{Q-values, the expected return at state $s$ \\after taking action $a$ following policy $\mu$ or $\pi$}\\\hline
		$Q_w(s,a)$ & estimated Q function \\\hline
		\tabincell{c}{$V^{\mu_{\theta}}(s)$,\\$V^{\pi}(s)$ }& \tabincell{c}{state value, the expected total \\discounted reward from state $s$} \\\hline
		%&\tabincell{c}{ $\alpha_{R}$,\\ $\alpha_{S},\alpha_A$}&positive learning rates, in $[0,1]$ \\\hline 
		$\epsilon$&\tabincell{c}{in $[0,1]$, the weighted parameter to balance RL and SL} \\\hline
	\end{tabular}
	%\vspace{-1.5em}
\end{table}

\subsection{Model Preliminaries}\label{sec2}
Q-learning~\cite{watkins1992q} is an off-policy learning scheme that finds a greedy policy $\mu(s)=arg max_aQ^\mu(s,a)$,
where $Q^{\mu}(s,a)$ denotes action-value or Q value and is used in a small discrete action space.
For deterministic policy, the Q value can be calculated with dynamic programming as follows:
\begin{equation}\label{dqn1}
Q^\mu(s_t,a_t) = \mathbb{E}_{r_t,s_{t+1}\sim E}[r(s_t,a_t)+\gamma Q^\mu(s_{t+1},\mu_{t+1})],
\end{equation}
where $E$ indicates the environment. Deep Q network (DQN)~\cite{mnih2015human} utilizes deep learning to estimate a non-linear Q function $Q_w(s,a)$
parameterized by $w$. The strategy of \emph{replay buffer} is adopted to gain independent and identical distribution of samples
for training. Moreover, DQN asynchronously updates a \emph{target network} $Q_w^{tar}$ to minimize the least square loss as follows:
\begin{equation}\label{dqn}
\begin{split}
L(w) = \mathbb{E}_{r_t,s_t\sim E}[(Q_w(s_t,a_t)-y^t)^2],\\
y^t = r(s_t,a_t)+ \gamma Q_w^{tar}(s_{t+1},\mu(s_{t+1})).
\end{split}
\end{equation}

Policy gradient is employed to handle continuous or high dimensional actions.
To estimate the parameter $\theta$ of $\pi_\theta$, we maximize the expected return
from the start states $\mathbb{E}_{r_i,s_i\sim \pi_\theta}[R_1]= \mathbb{E}_{r_i,s_i\sim \pi_\theta}[V^{\pi_\theta}(s_1)]$, which is reformulated as:
$
\mathbb{E}_{r_i,s_i\sim \pi_\theta}[V^{\pi_\theta}(s_1)] =J(\pi_\theta)= \int_\mathcal{S}\rho^{\pi_\theta}(s)\int_\mathcal{A}\pi_\theta(a\vert s)r(s,a)\texttt{d}s\texttt{d}a
$, where $ V^{\pi_\theta}(s_1)$ is the state value of the start state.
$\rho^{\pi_\theta}(s')=\int_\mathcal{S}\sum^T_{t=1}\gamma^{t-1}p_1(s)p(s\rightarrow s',t,\pi_\theta)\texttt{d}s$
is discounted state distribution, where $p_1(s_1)$ is the initial state distribution and $p(s\rightarrow s',t,\pi_\theta)$
is the probability at state $s'$ after transition of t time steps from state $s$. Policy gradient learns the parameter $\theta$ by the gradient $\nabla_\theta J(\pi_\theta)$
which is calculated using the \emph{policy gradient theorem}~\cite{sutton2000policy}:
\begin{equation}\label{pg}
\begin{split}
\nabla_\theta J(\pi_\theta) = \int_\mathcal{S}\rho^{\pi_\theta}(s)\int_\mathcal{A}\nabla_\theta \pi_\theta(a\vert s)Q^{\pi_\theta}(s,a) \texttt{d}a\texttt{d}s \\  = \mathbb{E}_{s\sim \rho^{\pi\theta},a\in \pi_\theta}[\nabla_\theta log \pi_\theta(a\vert s)Q^{\pi_\theta}(s,a)],
\end{split}
\end{equation}
where the instantaneous reward $r(s,a)$ is replaced by the long-term value $Q^{\pi\theta}(s,a)$.

Actor-critic~\cite{konda2000actor} combines the advantages of Q-learning and policy gradient to achieve accelerated and stable learning.
It consists of two components: (1) an actor to optimize the policy $\pi_\theta$ in the direction of gradient $\nabla_\theta J(\pi_\theta)$ using Equation \ref{pg}, and (2) a critic to estimate an action-value function $Q_w(s,a)$ with the parameter $w$ through Equation~\ref{dqn}.
Finally, we obtain the policy gradient denoted as
$\nabla_\theta J(\pi_\theta)  = \mathbb{E}_{s\sim \rho^{\pi_\theta},a\in \pi_\theta}[\nabla_\theta log \pi_\theta(a\vert s)Q_w(s,a)]$.

%With the conditions of (1) $Q_w(s,a)=\nabla_\theta log\pi_\theta(a\vert s)^\top w$; (2) $w$ is obtained by minimizing $\mathbb{E}_{s\sim \rho^\pi,a\sim \pi_{\theta}}[(Q_w(s,a)-Q^\pi(s,a))^2]$, 

In an off-policy setting, actor-critic estimates the value function of $\pi_\theta(a\vert s)$
by averaging the state distribution of behavior policy $\beta(a\vert s)$~\cite{degris2012model}. 
Instead of considering the stochastic policy $\pi_\theta(s\vert a)$, the deterministic policy gradient (DPG) theorem~\cite{silver2014deterministic} proves
that policy gradient framework can be extended to find deterministic off-policy $\mu_\theta(s)$, which is given as follows:
\begin{equation}\label{dpg}
%\vspace{-1pt}
\begin{split}
\nabla_\theta J_\beta(\mu_\theta) \approx \int_s\rho^\beta(s)\nabla_\theta\mu_\theta(s)Q^{\mu\theta}(s,a)\texttt{d}s\\= \mathbb{E}_{s\sim \rho^\beta}[\nabla_\theta\mu_\theta(s)\nabla_a Q^{\mu\theta}(s,a)\vert_{a=\mu_\theta(s)}].
\end{split}
\end{equation}

Deep deterministic policy gradient (DDPG)~\cite{lillicrap2015continuous} 
adopts deep learning to learn the actor and critic in mini batches which store a replay buffer with tuples ($s_t, a_t, r_t, s_{t+1}$).
To ensure the stability of Q values, DDPG uses a similar idea as the \emph{target network} of DQN to copy the actor and critic networks as
$\mu_\theta^{tar}(s)$ and $Q^{tar}_w(s,a)$. Instead of directly copying the weights, DDPG uses a ``soft'' target update:
\begin{equation}\label{soft_uppdate}
\theta^{tar} \leftarrow \tau\theta + (1-\tau)\theta^{tar},~w^{tar} \leftarrow \tau w  + (1-\tau)w^{tar},\quad \texttt{where} \quad \tau \ll 1.
\end{equation}

\section{SRL-RNN Architecture}\label{srl}
This section begins with a brief overview of our approach. After that, we introduce the components of SRL-RNN and the learning algorithm in detail. 
\subsection{Overview}
% Reinforcement learning learns a policy $\mu(s\vert \theta_{RL})$ by utilizing evaluation signals to maximize cumulated rewards in a sequential dynamic system. Supervised learning learns a policy $\mu(s\vert \theta_{SL})$ by minimizing the error signals from labeled training examples provided by knowledgable supervisors.
%Figure \ref{fig:model} shows our version of supervised reinforcement learning architecture (SRL-RNN).
The goal of our task is to learn a policy $\mu_{\theta}(s)$ to recommend tailored treatments given the dynamic states of patients. SL learns the policy $\mu_{ \theta}(s)$ by matching the indicator signals, which guarantees a standard and safe performance. But the ``good'' treatment strategy is unclear and the original goal of clinical decisions also tries to optimize the outcome of patients. RL learns the policy $\mu_{\theta}(s)$ by maximizing evaluation signals in a sequential dynamic system which reflects the clinical facts and can infer the optimal policy based on non-optimal strategies. But without supervision, RL may produce unacceptable medications which give high risk. Intuitively, the indicator signal and evaluation signal play complementary roles (demonstrated in Section~\ref{reasult}). Thus, SRL-RNN is proposed to combine these two signals at the same time, where the cumulated reward $r_t$ is the evaluation signal and the prescription $\hat{a}_t$ of doctors from the unknown policy $\hat{\mu}(s)$ is the indicator signal. 

%complementary \cite{barto1994reinforcement}
Figure \ref{fig:model} shows our proposed supervised reinforcement learning architecture (SRL-RNN),
which consists of three core networks: Actor ($Actor_{target}$), Critic ($Critic_{target}$), and LSTM. 
The actor network recommends the time-varying medications according to the dynamic states of patients,
where a supervisor of doctors' decisions provides the indicator signal to ensure safe actions and 
leverages the knowledge of doctors to accelerate learning process.
The critic network estimates the action value associated with the actor network to encourage or discourage the recommended treatments.
Due to the lack of fully observed states in the real world, LSTM is used to extend SRL-RNN to handle POMDP
by summarizing the entire historical observations to capture a more complete observations.

\subsection{Actor Network Update}
The actor network learns a policy $\mu_{\theta}(s)$ parameterized by $\theta$ to predict the time-varying treatments for patients, where the input is $s_t$ and the output is the prescription $a_t$ recommended by $\mu_{\theta}(s)$. We employ reinforcement learning and supervised learning to optimize the parameter $\theta$ jointly. By combining the two learning tasks together, we maximize the following objective function: 
\begin{equation}
\begin{split}
J(\theta) = (1-\epsilon) J_{RL}(\theta)+\epsilon (-J_{SL}(\theta)),
\end{split}
\end{equation}
where $J_{RL}(\theta)$ is the objective function of RL task (in Equation \ref{ob_rl}) which tries to maximize the expected return, $J_{SL}(\theta)$ is the objective function of SL task (in Equation \ref{ob_sl}) which tries to minimize the difference from doctors' prescriptions, and $\epsilon$ is a weight parameter to trade off the reinforcement learning and supervised learning tasks. Intuitively, our objective function aims to predicting mediations
which give both high expected returns and low errors.
Mathematically, the parameter $\theta$ of the learned policy is updated by gradient ascent as follows:
%\vspace{-3pt}
\begin{equation}
%\vspace{-3pt}
\label{sl_rl}
\theta = \theta + \alpha((1-\epsilon) \nabla J_{RL}(\theta)+\epsilon(- \nabla J_{SL}( \theta))),
\end{equation}
where $\alpha \in [0,1]$ is a positive learning rate, and $\nabla J_{RL}(\theta)$ and $\nabla J_{SL}( \theta)$ are acquired by the RL and SL tasks, respectively.
%where ${a^*}\in argmax_{a}(J(\theta^{RL})-J(\theta^{SL})$. 
%Thus $\theta$ is updated in the direction of gradient of RL task $\nabla J(\theta_{a^*}^{RL})$ while in the opposite direction of gradient of SL task $\nabla J(\theta_{a^*}^{SL})$ .

For the reinforcement learning task, we seek to learn the policy $\mu_{\theta}(s)$ by maximizing the state value of $\mu_{\theta}(s)$ averaging over the state distribution of the behaviors of doctors:
\begin{equation}
\label{ob_rl}
\begin{split}
J_{RL}(\theta) = \mathbb{E}_{s\sim \rho^{\hat{\mu}(s)}}[V^{\mu_{\theta}}(s)]
=\mathbb{E}_{s\sim \rho^{\hat{\mu}(s)}}[Q_w(s,\mu_{\theta}(s))].
\end{split}
\end{equation}
Let the parameter $\theta$ in RL task be the weights of a neural network.
$\theta$ is updated by gradient ascent, $\theta^{t+1} = \theta^t +\alpha (1-\epsilon)\nabla J_{RL}(\theta^{t})$, where $\nabla J_{RL}(\theta^{t})$ is calculated by the deterministic off-policy gradient using Equation \ref{dpg}:
\begin{equation}
\label{rl}
\begin{split}
\nabla J_{RL}(\theta^{t}) \approx  \mathbb{E}_{s_t\sim \rho^{\hat{\mu}(s)}}[\nabla_{\theta} \mu_{\theta}(s)\vert_{ s=s_t}\nabla_a Q_w(s,a)\vert_{s=s_t,a=\mu_{\theta}(s_t)}].
\end{split}
\end{equation}

Let $a_t$ be the treatment recommended by $\mu_{\theta}(s_t)$. $\nabla_a Q_w(s,a)$ is obtained by the chain rule,
which is used to tell the medications predicted by the actor are ``good'' or ``bad''.
When the gradient $\nabla_aQ_w(s_t,a_t)$ is positive, the policy $\mu_{\theta}(s)$ will be pushed to be closer to $a_t$.
Otherwise, $\mu_{\theta}(s)$ will be pushed away from $a_t$. $\nabla_{\theta} \mu_{\theta}(s)$ is a Jacobian matrix where each column is the gradient $\nabla_{\theta} [\mu_{\theta}(s)]_k$ of $k$-th medication of $\mu_{\theta}(s)$ with respect to $\theta$.

For the supervised learning task, we try to learn the policy $\mu_{\theta}(s)$ through minimizing the difference between treatments predicted by $\mu_{\theta}(s)$ and prescriptions given from the doctor's policy  $\hat{\mu}(s)$, using the cross entropy loss:
\begin{equation}
\begin{split}
\label{ob_sl}
J_{SL}({\theta^t}) = \mathbb{E}_{s\sim \rho^{\hat{\mu}(s)}}[-\frac{1}{K}\sum^K_{k=1}\hat{a}_{t,k}log\mu_{\theta}^k(s)+\\ (1-\hat{a}_{t,k})log(1-\mu_{\theta}^k(s)]\vert_{s=s_t},
\end{split}
\end{equation}
where $K$ indicates the number of medications or medication categories,
and $\hat{a}_{t,k}\in\{0,1\}$ denotes whether the doctor chooses $k$-th medication at time step $t$
and $\mu_{\theta}^k(s)$ is the probability of the $k$-th medication predicted by $\mu_{\theta}(s)$.
The parameter $\theta$ in SL task is updated by the gradient descent
$\theta^{t+1} = \theta^t -\alpha \epsilon\nabla J_{SL}(\theta^{t})$ where $ \nabla J_{SL}(\theta^{t})$ is derived by the chain rule as follows:
\begin{equation}
\label{sl}
\nabla J_{SL}(\theta^{t}) = \mathbb{E}_{s\sim \rho^{\hat{\mu}(s)}}[-\phi \nabla_{\theta} \mu_{\theta}(s\vert\theta^{t})\vert_{ s=s_t}],
\end{equation}
where $\phi=\frac{1}{K}\sum^K_{k=1}\frac{\hat{a}_{t,k}-{a}_{t,k}^{SL}}{(1-{a}_{t,k}^{SL}){a}_{t,k}^{SL}}$.
Substituting Equation \ref{rl} and Equation \ref{sl} into Equation \ref{sl_rl} gives the final actor update:
\begin{equation}
\label{eq:epsilon}
\begin{split}
\theta = \theta + \alpha [(1-\epsilon)\nabla_a Q_w(s,a)\vert_{s=s_t,a=\mu_{\theta}(s_t)}+\epsilon \phi]\nabla_{\theta} \mu_{\theta}(s)\vert_{ s=s_t}.
\end{split}
\end{equation}

The above formulas show that we can use $\epsilon$ to trade off the RL and SL tasks,
where $\epsilon$ is a hyper parameter in this paper
and the effect of $\epsilon$ is shown in Figure~\ref{fig:epsilon}.

%between utilizing evaluation signal $\nabla_a Q^\mu(s,a)\nabla_{\theta}\mu_{\theta}(s)$ and indicator signal $\phi\nabla_{\theta} \mu_{\theta}(s)$. %Thus, SRL equals supervised learning task when $\epsilon=0$, which prefers the actions same as doctor. SRL equals reinforcement learning task when $\epsilon=1$, which prefers the actions with high Q-values.

\subsection{Critic Network Update}
The critic network is jointly learned with the actor network, where the inputs are the states of patients,
doctor prescriptions, actor outputs, and rewards. We can do this due to the critic network is only needed for guiding the actor during training,
while only the actor network is required at test time. The critic network uses a neural network to learn the action-value function $Q_{w}(s, a)$
which is used to update the parameters of the actor in the direction of performance improvement $\nabla_a Q_w(s,a)$.
The output of the critic network is the Q value $Q(s_t,a_t)$ of state $s_t$
after taking action $a_t$, where $Q(s_t,a_t)$ is predicted by the Q-function $Q_{w}(s,a)$. 

The update formula for parameter $w$ of $Q_{w}(s,a)$ is similar to DQN which is specified in Equation \ref{dqn1} and \ref{dqn}:
\begin{equation}
\begin{split}
J(w) = \mathbb{E}_{r_t,s_t\sim \rho^{\hat{\mu}(s)}}[(Q_w(s_t,\hat{a}_t)-y^t)^2]\\
y^t = r(s_t,\hat{a}_t)+ \gamma Q_w^{tar}(s_{t+1},\mu_{\theta}(s_{t+1})),
\end{split}
\end{equation}
where $\mu_{t+1}(s_{t+1})$ are the medications recommend by the actor network. 
$w$ is updated by the gradient descent:
\begin{equation}
w = w-\alpha_Q (Q_w(s_t,\hat{a}_t)-y^t)\nabla_w Q_w(s_t,a_t),
\end{equation}
where $ \delta =(Q_w(s_t,\hat{a}_t)-y^t)$ is called Temporal Difference (TD) error which is used for learning the Q-function.

\subsection{Recurrent SRL}\label{srl-recurrent}
In the previous section, we assume the state $s_t$ of a patient is fully observed.
In fact, we are always unable to obtain the full states of the patient. 
Here we reformulate the environment of SRL-RNN as POMDP.
POMDP can be described as a 4-tuple ($\mathcal{S},\mathcal{A},\mathcal{R},\mathcal{O}$),
where $\mathcal{O}$ are observations.
We obtain the observation $o\in \mathcal{O} $ directly which conditions on $p(o_t\vert s_t)$, with the not fully observable state $s_t$. %Thus, the treatment trajectories of patients $\tau=(s_0,o_0,a_0,s_1,...)$ is generated by policy $\mu$: $p(s_0)p(o_0\vert s_0)\mu(c_0)...$. 
LSTM has been verified to improve performance on POMDP by summarizing entire historical observations $c_t=f(o_1,o_2,...,o_t)$ when using policy gradient~\cite{wierstra2007solving} and Q-learning~\cite{hausknecht2015deep}. In LSTM, the hidden units ${c}_t$ is to encapsulate the entire observation history $o_{1:t}$. %In principle, it is better to capture complete observations. 
In order to capture the dependence of the longitudinal and temporal records of the patient, we employ a LSTM with SRL to represent historical observation $c_t$ for a more complete observation. 
%In order to capture the independence of the longitudinal and temporal records of patients, we employ a LSTM with SRL to  up to time-step $t$. LSTM is a powerful model to use hidden units ${c}_t$ to encapsulate the entire historical states $s_{1:t}$. In principle, it is better to capture complete states.

%The hidden unit ${c}_t$ is obtained by LSTM as follows:
%\begin{equation}
%f_t = \sigma(W_f[h_{t-1},s_t]+b_f);
%\end{equation}
%the input gate layer gives the input rate $i_t$ of new observation:
% \begin{equation}
%i_t = \sigma(W_i[h_{t-1},s_t]+b_i),
% \end{equation}
% \begin{equation}
%\hat{c}_t =tanh(W_c[h_{t-1},s_t]+b_c),
% \end{equation}
% where $\hat{c}_t$ indicates the new observation.
% Finally we obtain the summary observations $c_t$ by the weighted sum of old and new observations:
% \vspace{-3pt}
%  \begin{equation}
%   \vspace{-3pt}
%c_t = LSTM(c_{t-1},o_t)
% \end{equation}

The updates of the parameters of actor network and critic network are modified as follows:
\begin{equation}
\begin{split}
\theta = \theta + \alpha \sum_t([(1-\epsilon)\nabla_a Q_w(c,a)\vert_{c=c_t,a=\mu_{\theta}(c_t)} \\+\epsilon\phi]\nabla_{\theta} \mu_{\theta}(c)\vert_{ c=c_t}),
\end{split}
\end{equation}
\begin{equation}
%\vspace{-3pt}
w = w-\alpha \sum_t(Q_w(c_t,\hat{a}_t)-y^t)\nabla_w Q_w(c_t,a_t).
\end{equation}

\subsection{Algorithm}
Putting all the aforementioned components together, the learning algorithm of SRL-RNN is provided below. %We first store the sequences of doctors' prescription in the buffer $D$, then train SRL-RNN by sampling the mini batches of sequences.

\begin{algorithm}[!h]  
	
	\caption{SRL-RNN}
	\label{algorithem1}
	\begin{algorithmic}[1]
		\REQUIRE observations $O$, actions $A$, reward function $r$, \# of epochs $N$, weight parameter $\epsilon$,  \# of medications $K$;
		\STATE Store sequences ($o_1,a_1,r_1,...,o_T,a_T,r_T$) in buffer D;
		\STATE Initialize actor $\mu_\theta(c)$, target actor $\mu_\theta(c)^{tar}$, critic $Q_w(c,a)$, target critic $Q_w(c,a)^{tar}$, and TD error $\delta=0$; 
		\FOR{$n=0$ to $N$} 
		\STATE Sample ($o_1^i,a_1^i,r_1^i,...,o_T^i,a_T^i,r_T^i$) $i=1,..,I$ from D;
		\STATE $c_t^i \leftarrow f(o_1^i,o_2^i,...,o_t^i)$ given by LSTM
		\STATE $y_t^i  \leftarrow r_t^i+ \gamma Q^{tar}_w(c_t^{i+1},\mu_\theta^{tar}(c_t^{i+1}))$
		\STATE ${a}_t^i\leftarrow $ action given by  $\mu_\theta(c_t^i)$
		\STATE $\hat{a}_t^i\leftarrow $ action given by doctor $\hat{\mu}(c_t^i)$
		
		\STATE $\delta_t^i \leftarrow Q_w(c_t^i,\hat{a}_t^i)-y_t^i$
		\STATE $w \leftarrow w+\alpha\frac{1}{IT}\sum_i\sum_t\delta_t^i \nabla_w Q_w(c_t^i,a_t^i)$ \hfill $\vartriangleright$ update critic
		\STATE $ w^{tar} \leftarrow \tau w + (1-\tau)w^{tar}$   \hfill $\vartriangleright$ update target critic
		\STATE $\nabla_a Q_w(c_t^i,{a}_t^i)\leftarrow $ given by  $Q_w(c_t^i,a_t^i)$
		\STATE $\eta_t^i \leftarrow \frac{1}{K}\sum^K_{k=1}\frac{\hat{a}_{t,k}^i-{a}_{t,k}^i}{(1-{a}_{t,k}^i){a}_{t,k}^i}$
		\STATE $\theta \leftarrow \theta + \alpha \frac{1}{IT}\sum_i\sum_t[(1-\epsilon)\nabla_a Q_w(c_t^i,a_t^i)+\epsilon\eta_t^i]$\\$ \nabla_{\theta} \mu_{\theta}(c_t^i)$ \hfill $\vartriangleright$ update actor
		\STATE $ \theta^{tar} \leftarrow \tau\theta + (1-\tau)\theta^{tar}$\hfill $\vartriangleright$ update target actor
		\ENDFOR 
	\end{algorithmic}  
\end{algorithm}

\section{Experiments\label{exp}}
In this section, we present the quantitative and qualitative experiment results on MIMIC-3. 
%We first report the preprocessing steps for the dataset, followed by the evaluation metrics and comparison methods we adopted.

%Then we design the experiments to answer the questions below:

%\noindent\textbf{Q1. Performance of SRL-RNN:} 
%How much mortality rate could SRL-RNN reduce for patients in ICU? And
%how well does the policy generated by SRL-RNN match the prescriptions by clinicians?

%\noindent\textbf{Q2. Ablation study:}
%What are the contributions of different types of patient-specific information in different learning settings?  

%\noindent\textbf{Q3. Effectiveness and stability of policy:}
%What is the correlation between estimated mortality rate and expected return of the learned policy? 
%Are the mortality rates low when optimal policy and true prescriptions coincide?
%And does the policy learned in a stable manner?

%\noindent\textbf{Q4. Case studies:} What are the detailed prescriptions predicted by the optimal policy
%for randomly selected patients? And what are the efficacy of them?

\subsection{Dataset and Cohort}
The experiments are conducted on a large and publicly available dataset,
namely the Multiparameter Intelligent Monitoring in Intensive Care (MIMIC-3 v1.4) database~\cite{johnson2016mimic}. MIMIC-3 encompasses a very large population of patients compared to other public EHRs.
It contains hospital admissions of 43K patients in critical care units during 2001 and 2012,
involving 6,695 distinct diseases and 4,127 drugs. To ensure statistical significance, we extract the top 1,000 medications
and top 2,000 diseases (represented by ICD-9 codes) which cover 85.4\% of all medication records and 95.3\% of all diagnosis records, respectively. 
In order to experiment on different granularity of medications, we map the 1,000 medications into the third level of
ATC\footnote{http://www.whocc.no/atc/structure and principles/} (medication codes) using a public tool\footnote{https://www.nlm.nih.gov/research/umls/rxnorm/},
resulting 180 distinct ATC codes. Therefore, the action space size of the experiments
is 1,000 exact medications ($K=1000$) or 180 drug categories ($K=180$).

For each patient, we extract relevant physiological parameters with the suggestion of clinicians,
which include static variables and time-series variables. The static variables cover eight kinds of demographics: gender, age, weight, height, religion,
language, marital status, and ethnicity. The time-series variables contain lab values, vital signs, and output events,
such as diastolic blood pressure, fraction of inspiration O2, Glascow coma scale, blood glucose, systolic blood pressure, heart rate, pH, respiratory rate, blood oxygen saturation, body temperature, and urine output. These features correspond to the state $s$ in MDP or the observation $o$ in POMDP.
We impute the missing variable with k-nearest neighbors and remove admissions with more than 10 missing variables. Each hospital admission of a patient is regarded as a treatment plan. Time-series data in each treatment plan is divided into different units, each of which is set to 24 hours
since it is the median of the prescription frequency in MIMIC-3.
If several data points are in one unit, we use their average values instead.
Following~\cite{weng2017representation}, we remove patients less than 18 years old because of the special conditions of minors.
Finally, we obtain 22,865 hospital admissions, and randomly divide the dataset for training,
validation, and testing sets by the proportion of 80/10/10.
%The summary of the final obtained populations is shown in Table~\ref{tab:data}.

\subsection{Evaluation Metrics}\label{sec:evaluation-metric}
Evaluation methodology in treatment recommendation is still a chanllenge. Thus we try all evaluation metrics used in state-of-art methods to judge our model.

Following~\cite{weng2017representation,raghu2017deep}, we use the estimated in-hospital mortality rates
to measure whether policies would eventually reduce the patient mortality or not. 
Specifically, we discretize the learned Q-values of each test example into different units
with small intervals shown in the x-axis of Figure~\ref{fig:compare}.
Given an example denoting an admission of a patient,
if the patient died in hospital, all the Q-values belonging to this admission
are associated with a value of 1 as mortality and the corresponding units add up these values.
After scanning all test examples, the average estimated mortality rates for each unit are calculated, shown in y-axis of Figure~\ref{fig:compare}.
Based on these results, the mortality rates corresponding to the expected Q-values of different policies
are used as the measurements to denote the estimated in-hospital mortality (see details in~\cite{weng2017representation,raghu2017deep}).
Although the estimated mortality does not equal the
mortality in real clinical setting, it is a universal metric currently for computational testing.

Inspired by~\cite{Zhang2017LEAP}, we further utilize mean Jaccard coefficient to measure the degree of consistency
between prescriptions generated by different methods and those from doctors.
For a patient $p_i$ in the $t$-th day of ICU, let $\hat{U}_t^i$ be the medication set given by doctors and 
$U_t^i$ be the medication set recommended from learned policies.
The mean Jaccard is defined as $ \frac{1}{M}\sum_{i=1}^{M}\frac{1}{T_i} \sum_{t=1}^{T_i}\frac{\mid U_t^i\cap \hat{U}_t^i \mid}{\mid U_t^i\cup \hat{U}_t^i \mid }$,
where $M$ is the number of patients and $T_i$ is the number of ICU days for the patient $p_i$.

%We utilize these two metrics to evaluate SRL-RNN,
%where a mortality-expected-return function is to measure the reinforcement learning component
%and Jaccard is to measure the the supervised learning component.

\subsection{Comparison Methods}
All the adopted baselines models in the experiments are as follows, where BL, RL, and SD3Q are the alternatives of SRL-RNN we proposed.

\begin{itemize}[leftmargin=*]
	\item[-] \textbf{Popularity-20 (POP-20)}:
	POP-20 is a patten-based method, which chooses the top-K most co-occurring medications
	with the target diseases as prescriptions. We set $K=20$ for its best performance on the validation dataset.
	%Inspired by \cite{pmlr-v56-Choi16}, 
	
	\item[-] \textbf{Basic-LSTM (BL)}:
	BL uses LSTM to recommend the sequential medications based on the longitudinal and temporal records of patients.
	Inspired by Doctor-AI~\cite{pmlr-v56-Choi16}, BL fuses multi-sources of patient-specific information
	and considers each admission of a patient as a sequential treatment to satisfy the DTR setting.
	BL consists of a 1-layer MLP (M-1) to model diseases, a 1-layer MLP (M-2) to model static variables, and
	a 1-layer LSTM sequential model (L-1) to capture the time-series variables. 
	These outputs are finally concatenated to predict prescriptions at each time-step.
	
	\item[-] \textbf{Reward-LSTM (RL)}:
	RL has the same framework as BL, 
	except that it considers another signal, i.e., feedback of mortality, to learn a nontrivial policy. 
	The model involves three steps: (1) clustering the continuous states into discrete states, 
	(2) learning the $Q$-values using tabular Q-learning,
	(3) and training the model by simultaneously mimicking medications generated by doctors
	and maximizing cumulated reward of the policy.
	
	\item[-] \textbf{Dueling Double-Deep Q learning (D3Q)~\cite{raghu2017deep}}:
	D3Q is a reinforcement learning method which combines dueling Q, double Q, and deep Q together. D3Q regards a treatment plan as DTR.
	
	\item[-] \textbf{Supervised Dueling Double-Deep Q (SD3Q)}:
	Instead of separately learning the Q-values and policy as RL, SD3Q learns them jointly.
	SD3Q involves a D3Q architecture, where supervised learning is additionally adopted to revise the value function.
	
	\item[-] \textbf{Supervised Actor Critic (SAC)~\cite{Clouse1992A}}:
SAC uses the indicator signal to pre-train a ``guardian'' and then combines ``actor'' output and ``guardian'' output
	to send low-risk actions for robots. We transform it into a deep model for a fair comparison.
	
	\item[-] \textbf{LEAP~\cite{Zhang2017LEAP}}: LEAP leverages a MLP framework to train a multi-label model with the consideration of the dependence among medications. LEAP takes multiple diseases as input and multiple medications as output. Instead of considering each admission as a sequential treatment process, LEAP regards each admission as a static treatment setting.
	We aggregate the multiple prescriptions recommended by SRL-RNN as SRL-RNN (agg) for a fair comparison.
	
	\item[-] \textbf{LG~\cite{bajor2017predicting}}:
	LG takes diseases as input and adopts
	a 3-layer GRU model to predict the multiple medications.
	
	%\item[-] \textbf{SRL-RNN}:
	%This is the model proposed in this paper.
	%Instead of learning policy by modifying the value function as D3Q,
	%SRL-RNN involves an off-policy actor-critic architecture (DDPG) to directly modify the policy by
	%incorporating standard supervised learning. 
	%We aggregate the reasult of SRL-RNN  as SRL-RNN(agg) to compare with LEAP.
\end{itemize}

\begin{table}[!t]
	\caption{Performance comparison on test sets for prescription prediction. \emph{l-3 ATC} indicates the third level of ATC code and \emph{Medications} indicates the exact drugs.}
	\label{tab:all}
	%\vspace{-1.1em}
	\scalebox{0.95}{
		\begin{tabular}{cccccl}
			\toprule
			& \multicolumn{2}{c}{Estimated Mortality}& \multicolumn{2}{c}{Jaccard}\\
			\midrule
			\texttt{Granularity} &l-3 ATC&Medications&l-3 ATC&Medications \\
			\midrule
			\multicolumn{5}{c}{Dynamic treatment setting}\\\hline
			\texttt{LG}&0.226& 0.235 &0.436&0.356\\
			\texttt{BL}&0.217 &0.221 &0.512&0.376\\
			\texttt{RL}& 0.209&0.213 &0.524&0.378\\
			\texttt{D3Q}&0.203& 0.212 &0.109&0.064\\
			\texttt{SD3Q}&0.198& 0.201 &0.206&0.143\\
			\texttt{SAC}&0.196& 0.202 &0.517&0.363\\
			\texttt{SRL-RNN}&0.153& \textbf{0.157}&0.563 &\textbf{0.409}\\\hline
			\multicolumn{5}{c}{Static treatment setting}\\\hline
			\texttt{POP-20} &0.233&0.247&0.382& 0.263 \\
			\texttt{LEAP}&0.224&0.229&0.495& 0.364\\
			\texttt{SRL-RNN (agg)}&0.153& \textbf{0.157}&0.579 &\textbf{0.426}\\
			\bottomrule
		\end{tabular}
	}
	%\vspace{-1em}
\end{table}

\subsection{Result Analysis}\label{reasult}
%In this section, we design experiments to answer the aforementioned questions.
%The comparison methods are validated by a set of measurements. Mortality and Jaccard are used to measure the effectiveness and accuracy of the dynamic prescriptions. An ablation study is to evaluate different contributions of each factor in AC. RL outperforms BL by In addition, qualitatively measurements are used to assess the effectiveness of Q-value and treatment policy.

\begin{figure*}[ht]
	\centering
	%\vspace{-0.2em}
	\subfigure[all-demo-disease]{%
		\includegraphics[width=2in]{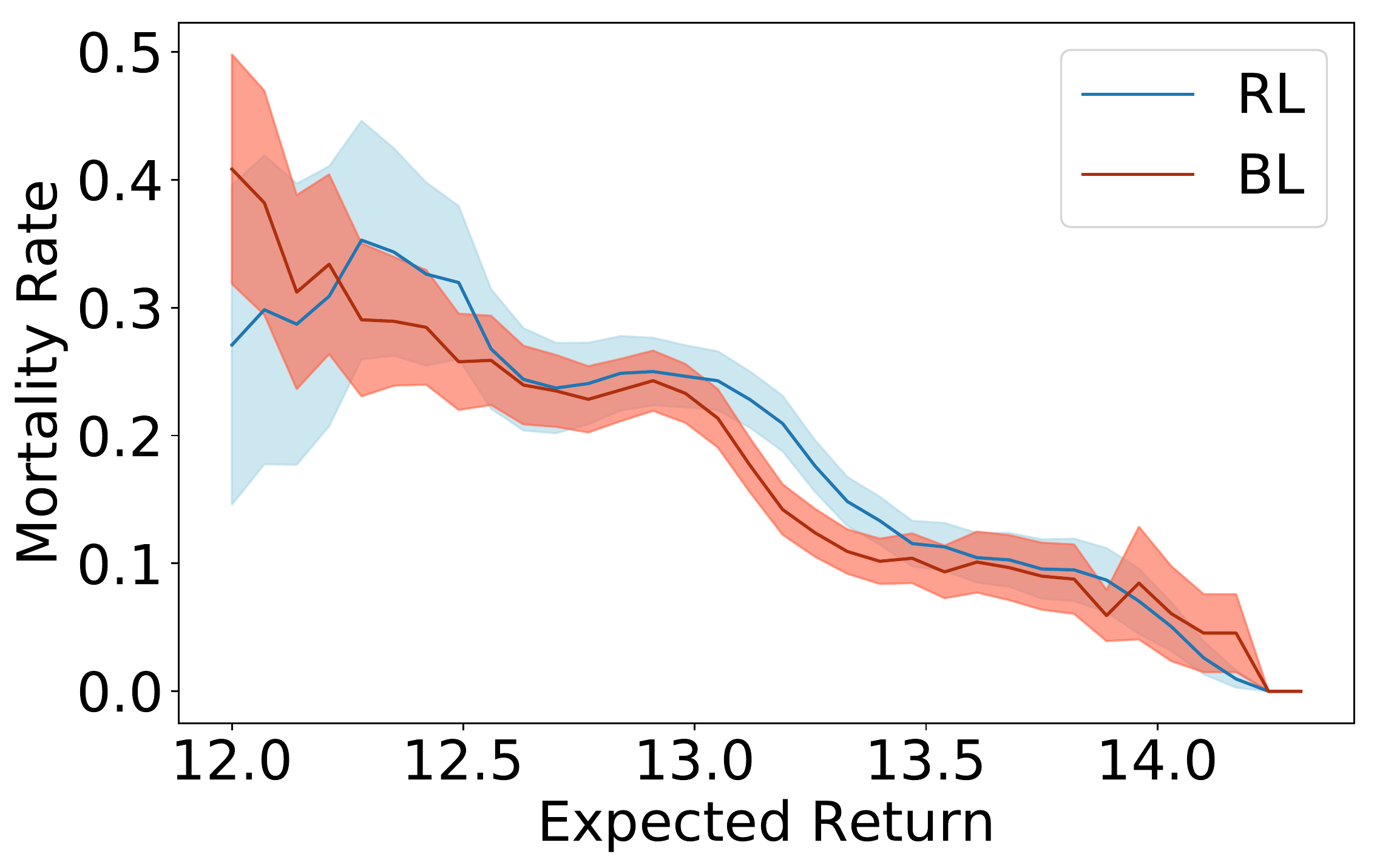}
		\label{fig:subf1}}
	%\vspace{-0.2em}
\quad
	\subfigure[all-demo]{%
		\includegraphics[width=2in]{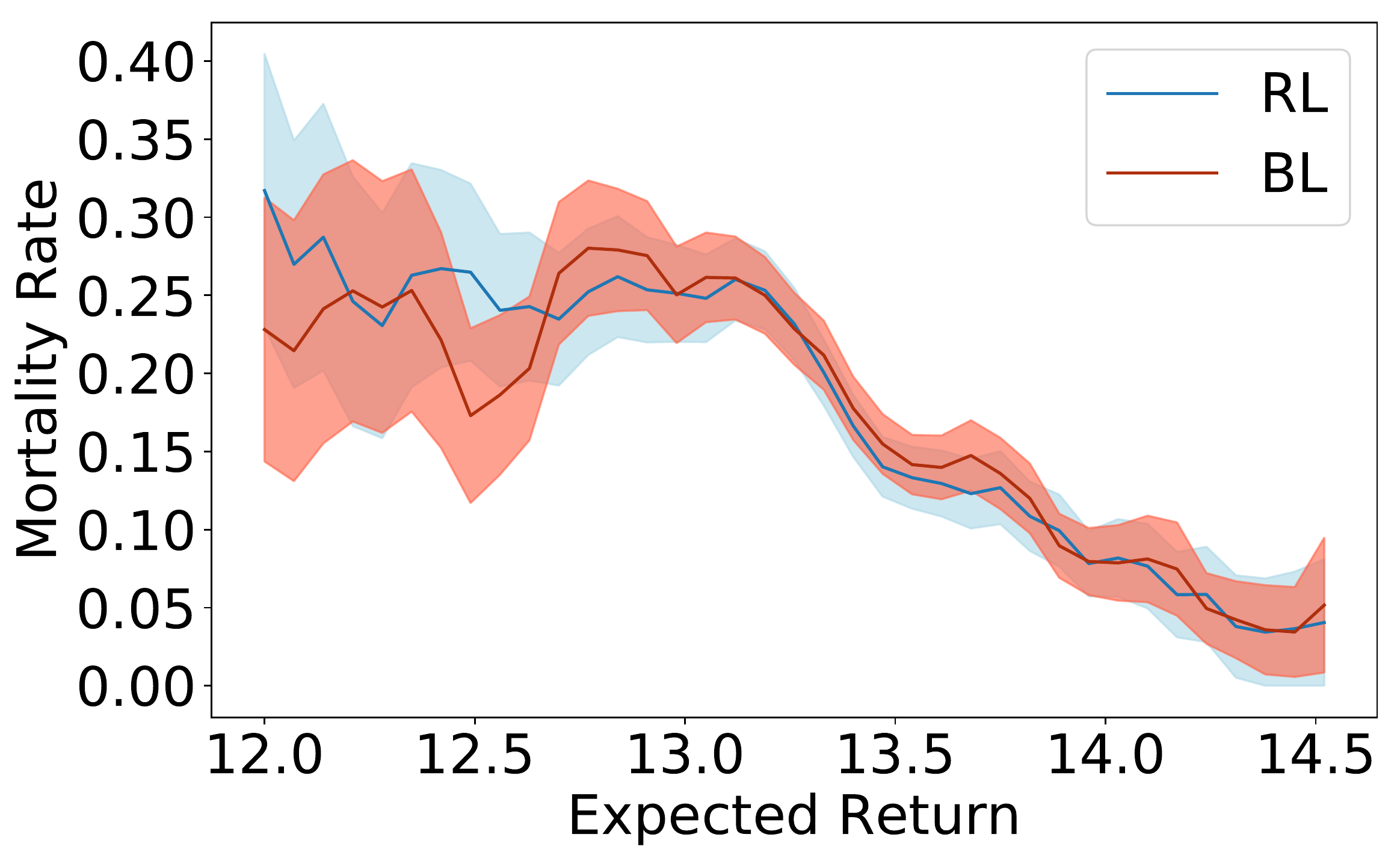}
		\label{fig:subfigure2}}
	%\vspace{-0.2em}
\quad
	\subfigure[all]{%
		\includegraphics[ width=2in]{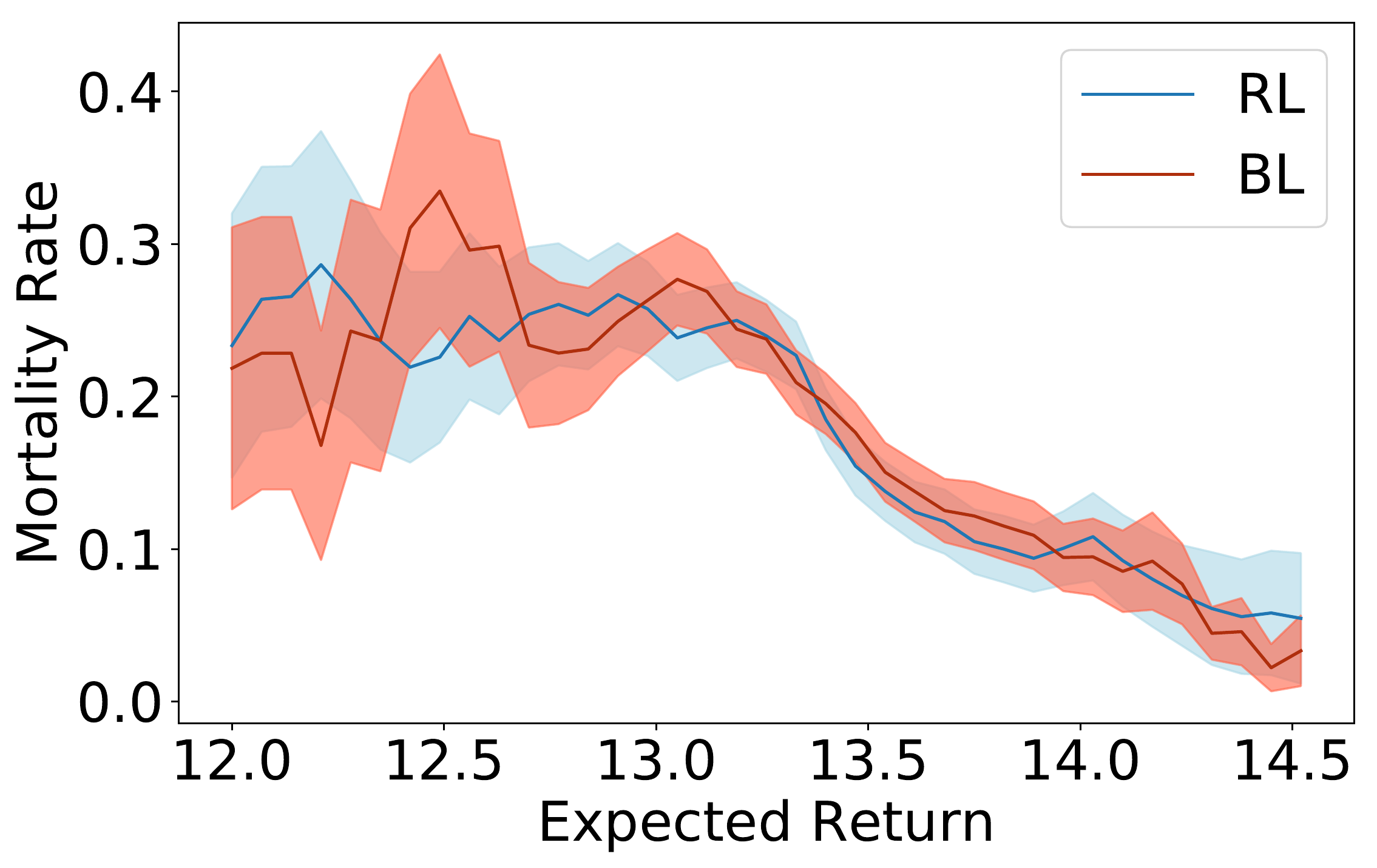}
		\label{fig:subfigure3}}
	\subfigure[all-demo-disease]{%
		\includegraphics[width=2in]{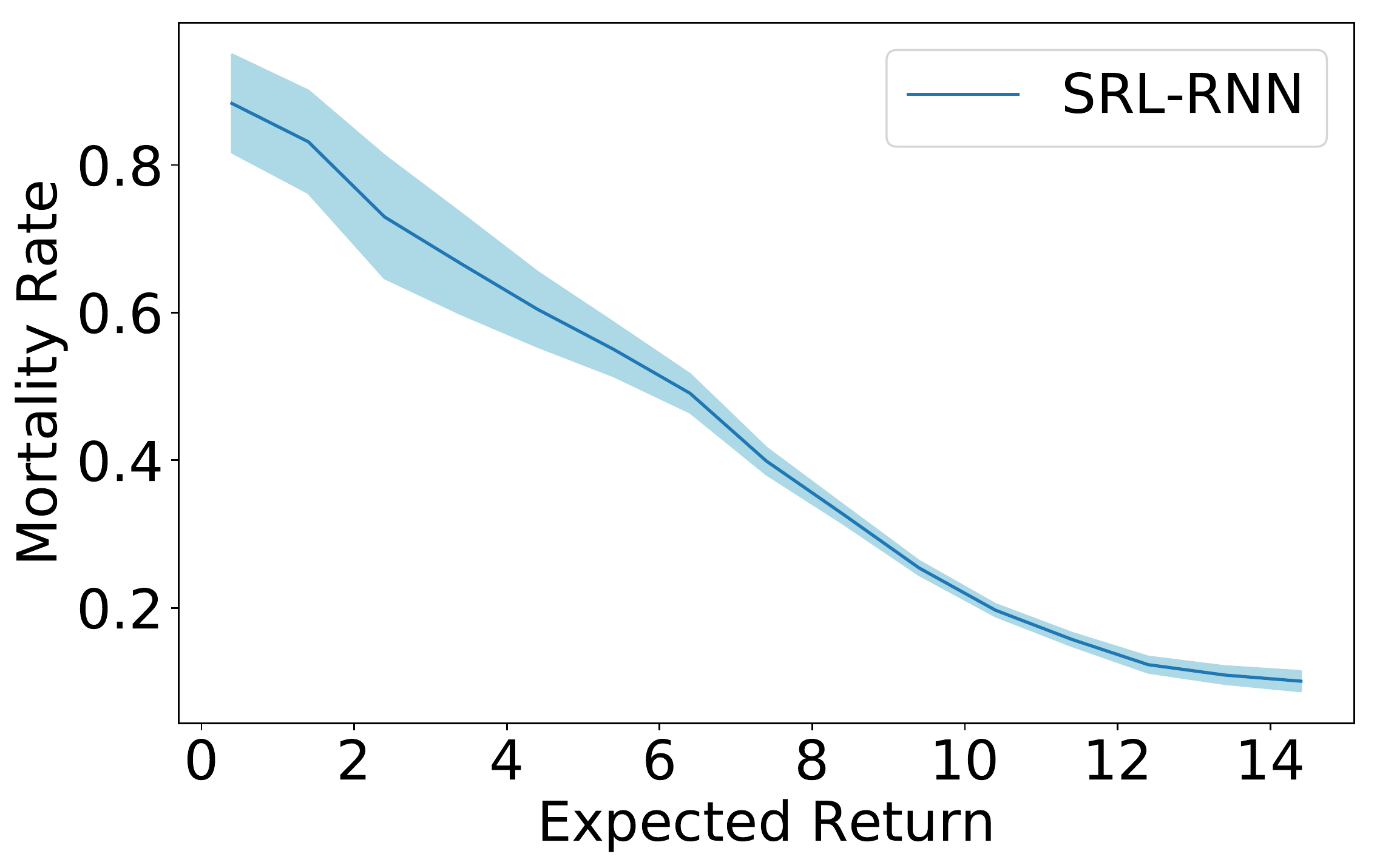}
		\label{fig:subfigure4}}
\quad
	\subfigure[all-demo]{%
		\includegraphics[ width=2in]{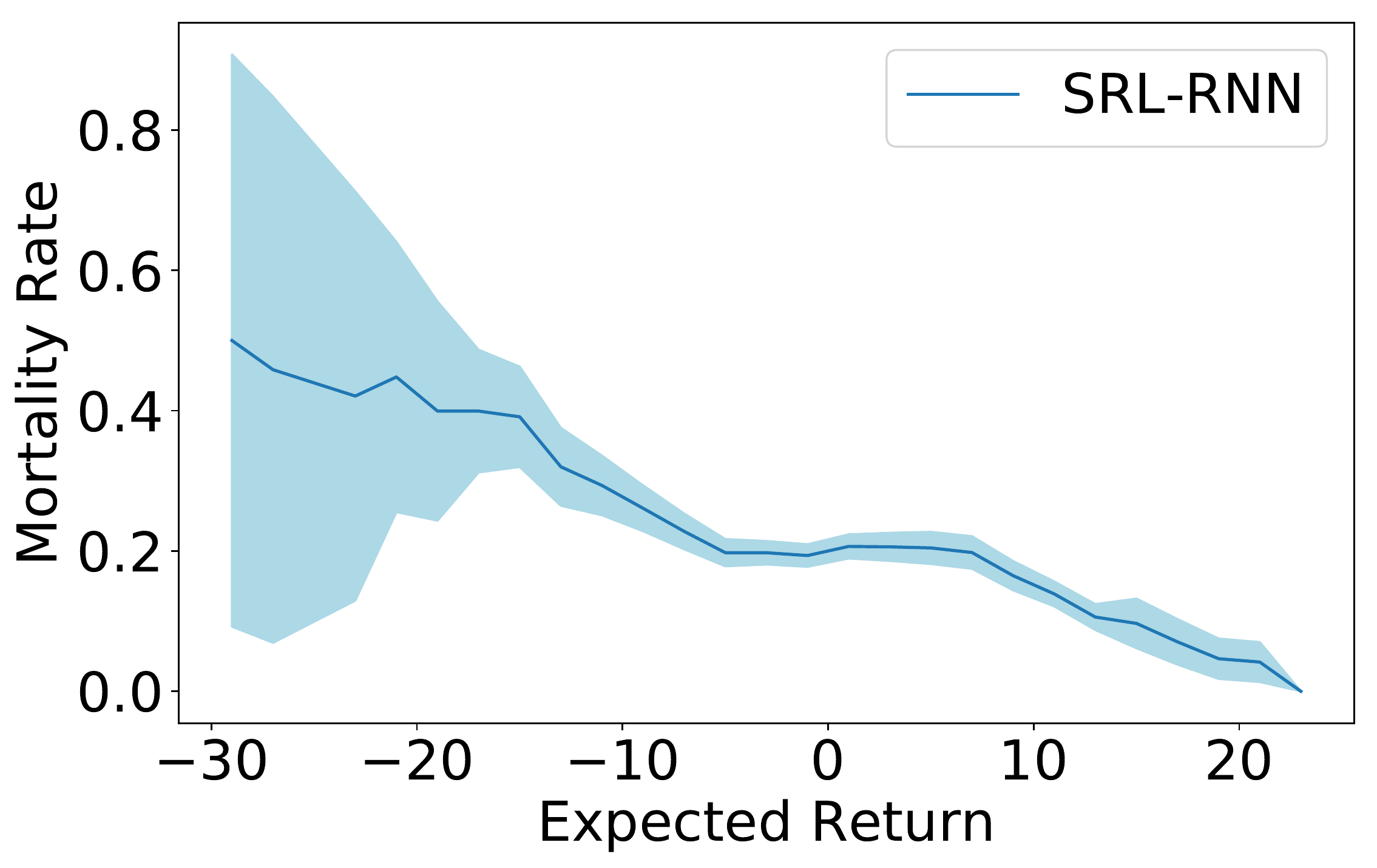}
		\label{fig:subfigure5}}
\quad
	\subfigure[all]{%
		\includegraphics[ width=2in]{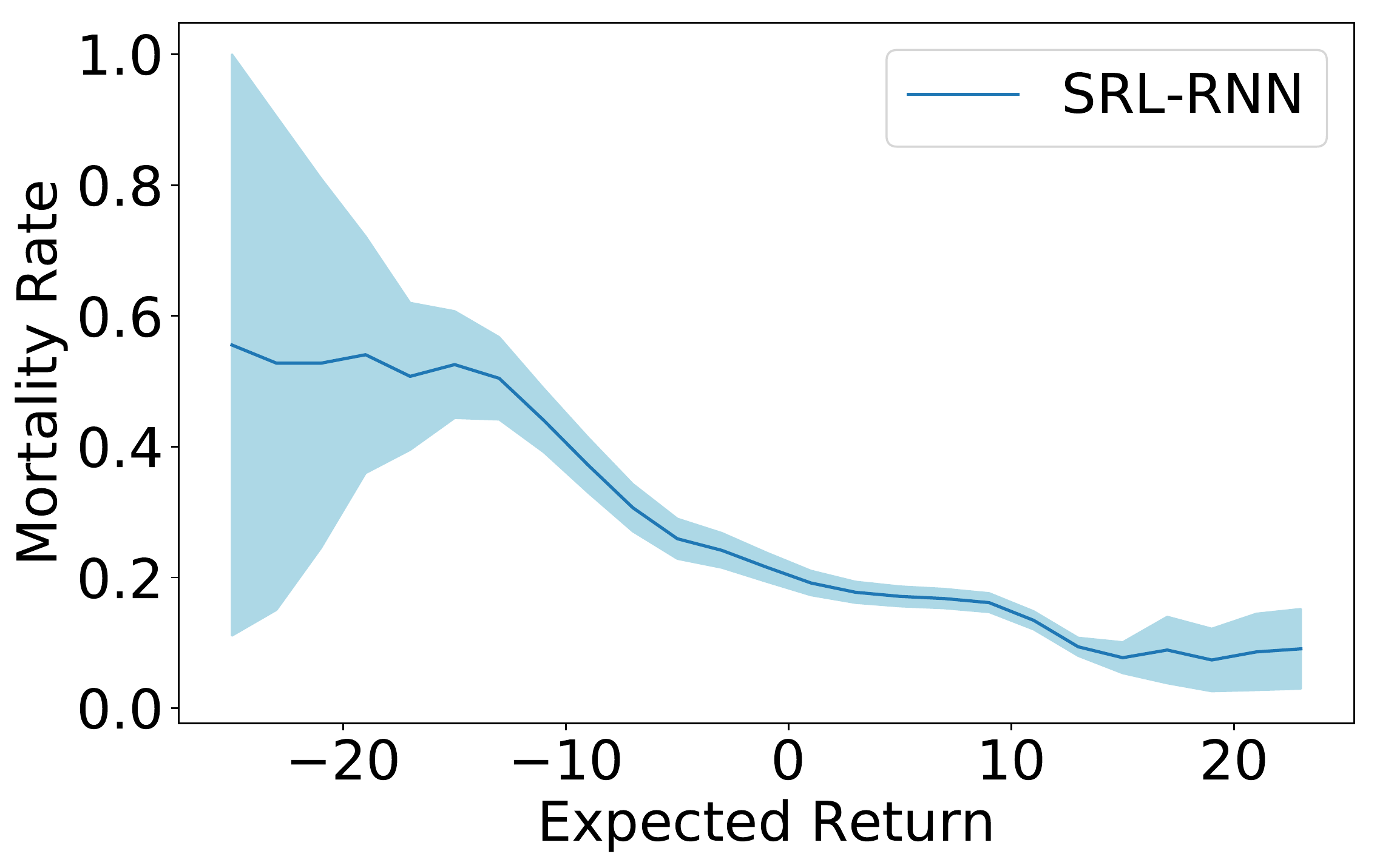}
		\label{fig:subfigure6}}
	%\vspace{-1em}
	%
	\caption{Mortality-expected-return curve computed by the learned policies. }
	\label{fig:figure}
%	\vspace{-5pt}
	\label{fig:compare}
\end{figure*}

\noindent \textbf{Model comparisons.}
Table \ref{tab:all} depicts the mortality rates and Jaccard scores for all the adopted models on MIMIC-3.
%BL outperforms LEAP and LG both on Jaccard and mortality. This result shows that considering rich patient-specific information
%might obtain better state representations of patients. 
By first comparing the results of LG, BL, and RL, 
we find that RL outperforms both BL and LG,
showing that incorporating the evaluation signal into supervised learning
can indeed improve the results.
We then compare SD3Q with its simplified version D3Q.
The better performance of SD3Q indicates the knowledgeable supervision guarantees a standard performance of the learned policy. 
In the static treatment setting, LEAP improves the basic method POP-20 by a large margin,
which demonstrates that capturing the relations between multiple diseases and multiple medications
is beneficial for better prescriptions.

Finally, our proposed model SRL-RNN performs significantly better than all the adopted baselines,
both in the dynamic treatment setting and static treatment setting.
The reasons are: 1) SRL-RNN regards the treatment recommendation as a sequential decision process,
reflecting the clinical practice (compared with LEAP and POP-20), and utilizes evaluation signal to infer an optimal treatment (compared with LG and BL);
2) SRL-RNN considers the prescriptions of doctors as supervision information to learn a robust policy (compared with D3Q), 
and applies the off-policy actor-critic framework to handle complex relations of medications, diseases, and individual characteristics (compared with SD3Q);
3) SRL-RNN integrates supervised learning and reinforcement learning in an end-to-end learning fashion
for sharing the information between evaluation signal and indicator signal (compared with RL and SAC);
and 4) SRL-RNN adopts RNN to solve POMDP by obtaining the representations of the entire historical observations. 

\noindent \textbf{Ablation study.}
The different contributions of the three types of features are reported in this part.
To be specific, we progressively add the patient-specific information, i.e., time series variables, diseases, and demographics,
into the selected models. As shown in Table \ref{tab:ablation}, the Jaccard scores of the three methods monotonically increase.
In addition, the estimated mortality rates of SRL-RNN monotonically decrease.
However, the estimated mortality variations of BL and RL are not the same as that of SLR-RNN. 
It might be due to the fact that with constant Q-values learned by tabular Q-learning,
learning a robust policy is a little hard.

\begin{table}{\centering}
	\caption{Ablation study of the features. \emph{all} indicates all the features: demographic, diseases and time-series variables. The symbol ``-'' stands for ``subtracting''.}
	\label{tab:ablation}
	%\vspace{-1.1em}
	\begin{tabular}{ccccl}
		\toprule
		Method& Feature&Estimated Mortality&Jaccard\\
		\midrule
		\multirow{3}{*}{BL} &\texttt{all-demo-disease}& 0.242 &0.323\\
		& \texttt{all-demo}&0.212& 0.360\\
		&\texttt{all}& 0.221 &0.376\\\hline
		\multirow{3}{*}{RL}   &
		\texttt{all-demo-disease} &0.184& 0.332 \\
		& \texttt{all-demo}& 0.203 &0.371\\
		& \texttt{all}&0.213& 0.378\\\hline
		\multirow{3}{*}{SRL-RNN}   &
		\texttt{all-demo-disease} &0.173& 0.362 \\
		& \texttt{all-demo}& 0.162 &0.403\\
		& \texttt{all}&0.157& 0.409\\
		\bottomrule
	\end{tabular}
	%\vspace{-1.5em}
\end{table}

\noindent \textbf{Effectiveness and stability of policy.}
The relations between expected returns and mortality rates are shown in Figure~\ref{fig:compare}.
We observe SRL-RNN has a more clear negative correlation between expected returns and mortality rates
than BL and RL. The reason might be that BL ignores the evaluation signal
while RL discretizes the continuous states, incurring information loss.

\begin{figure}
	\includegraphics[width=3in]{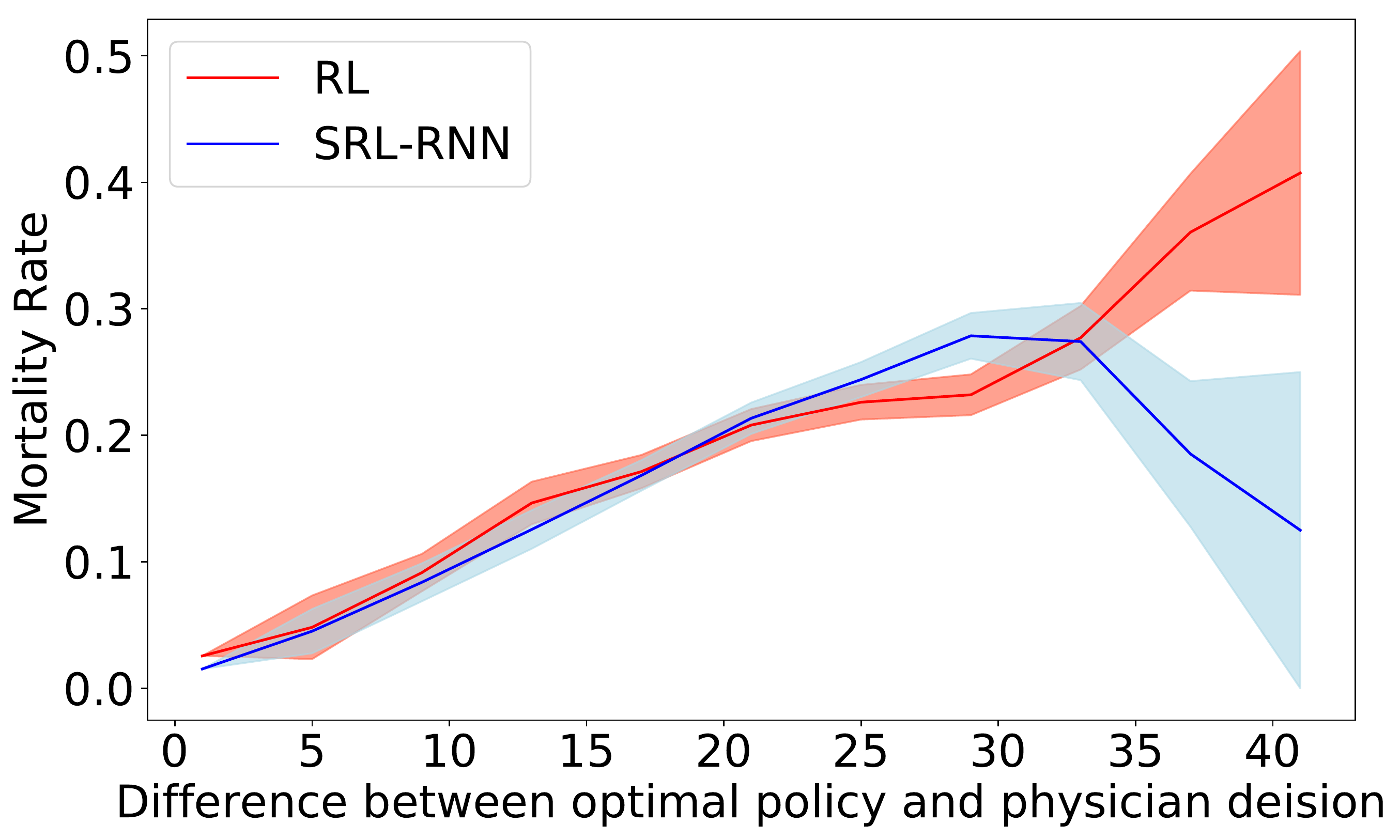}
	%\vspace{-1.5em}
	\caption{Comparison of how observed mortality rates (y-axis) vary with the difference between the prescriptions generated by the optimal policy and the prescriptions administered by doctors (x-axis).} \label{fig:diff}
	%	\vspace{-1em}
\end{figure}

Figure~\ref{fig:diff} shows how the observed mortality changes with
the difference between the learned policies (by RL and SRL-RNN) and doctors' prescriptions.
Let $D_t^i$ be the treatment difference for patient $p_i$ in the $t$-th day, and $K$ is the number of candidate classes of medications.
We calculate the difference by $D_t^i =  \sum_{k=1}^{K}{\mid U_{t,k}^i -\hat{U}_{t,k}^i \mid}$.
When the difference is minimum, we obtain the lowest mortality rates of 0.021 and 0.016 for RL and SRL-RNN, respectively.
This phenomenon shows that SRL-RNN and RL can both learn good policies while 
SRL-RNN slightly outperforms RL for its lower mortality rate.

\begin{figure}[ht]
	\includegraphics[width=3in]{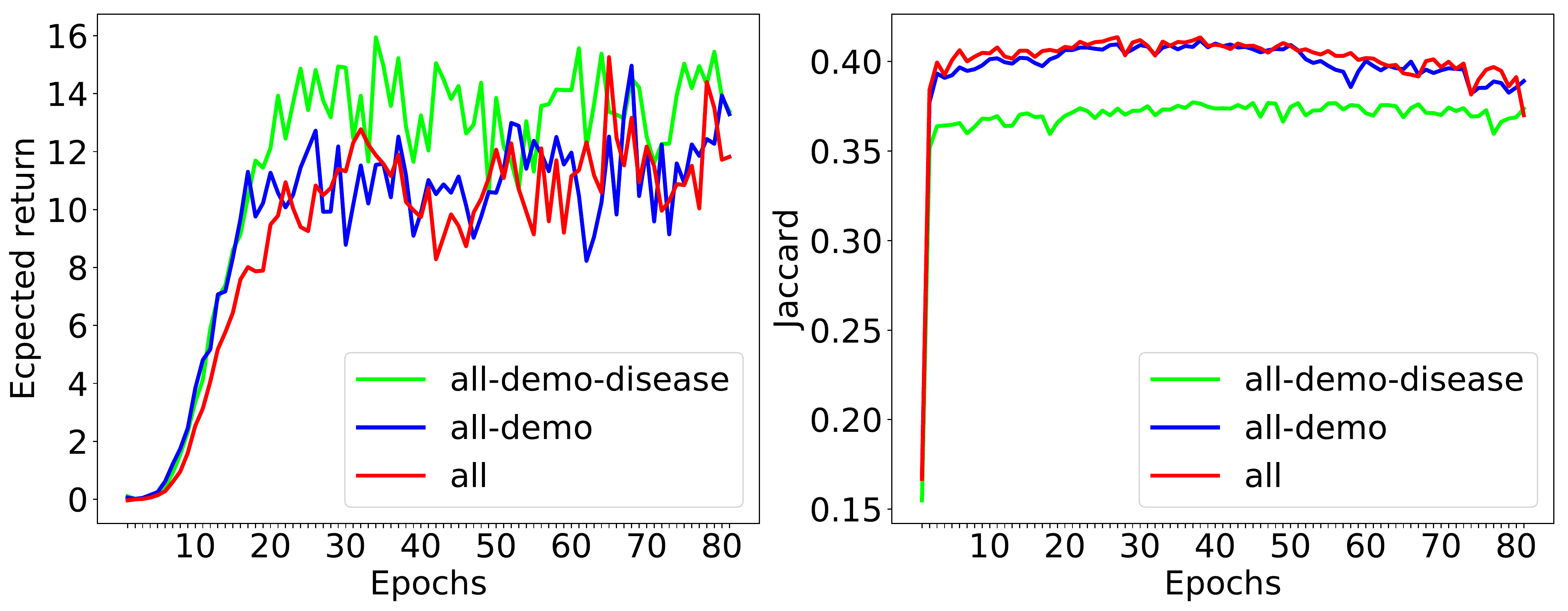}
	%\vspace{-1.5em}
	\caption{Expected return and Jaccard in different learning epochs with different features.}
	%\vspace{-0.5em}
	\label{fig:jar_mor}
\end{figure}

Figure~\ref{fig:jar_mor} presents the expected return and Jaccard scores obtained in each learning epoch of SRL-RNN with different features.
Notably, SRL-RNN is able to combine reinforcement learning and supervised learning to obtain the optimal policy in a stable manner.
We can see Jaccard is learned faster than Q-values, indicating that leaning Q-values might need rich trajectories.

%\begin{figure}[ht]
%\centering
%\subfigure[Expected Return]{%
% \includegraphics[width=3in]{qv.pdf}
%  \label{fig:subfigure1}}
%\quad
%\subfigure[Jaccard]{%
%  \includegraphics[width=3in]{jac.pdf}
%  \label{fig:subfigure2}}

%
%\caption{Expected Return and Jaccard obtained by each recommendation epochs on different states granularity.}
%\label{fig:figure}
%\end{figure}

%\begin{figure}[ht]
%\centering
%\subfigure[patient 1]{%
%  \includegraphics[height=1.2in, width=1.5in]{case2.pdf}
% \label{fig:subfigure1}}
%\quad
%\subfigure[patient 2]{%
%  \includegraphics[height=1.2in, width=1.5in]{case1.pdf}
%  \label{fig:subfigure2}}

%
%\caption{Dynamic states curve of two patients, where patient 1 is non-survivor and patient 2 is survivor. }
%\label{fig:figure}
%\end{figure}

\noindent \textbf{Case studies.}
Table~\ref{tab:case} shows the prescriptions generated by different models for two patients in different ICU days.
For the first patient who is finally dead, LG recommends much more medications which seem to be non-informative.
LEAP generates the same medications for patients in different ICU days, without considering the change of patient states.
An interesting observation is that the prescriptions recommended by SRL-RNN are much different from doctors',
especially for the stronger tranquilizers such as Acetaminophen and Morphine Sulfate.
Actually, for an ICU patient with a severe trauma, it is important to give full sedation in early treatment.
Considering the second surviving patient, the similarity between the prescriptions of SRL-RNN and doctor becomes larger,
indicating the rationality of SRL-RNN.
Particularly, all the models except SRL-RNN recommend redundant medications in different days, while
the prescriptions of doctors are significantly different.

\begin{figure}[h]
	
	\centering
	\includegraphics[width=3in]{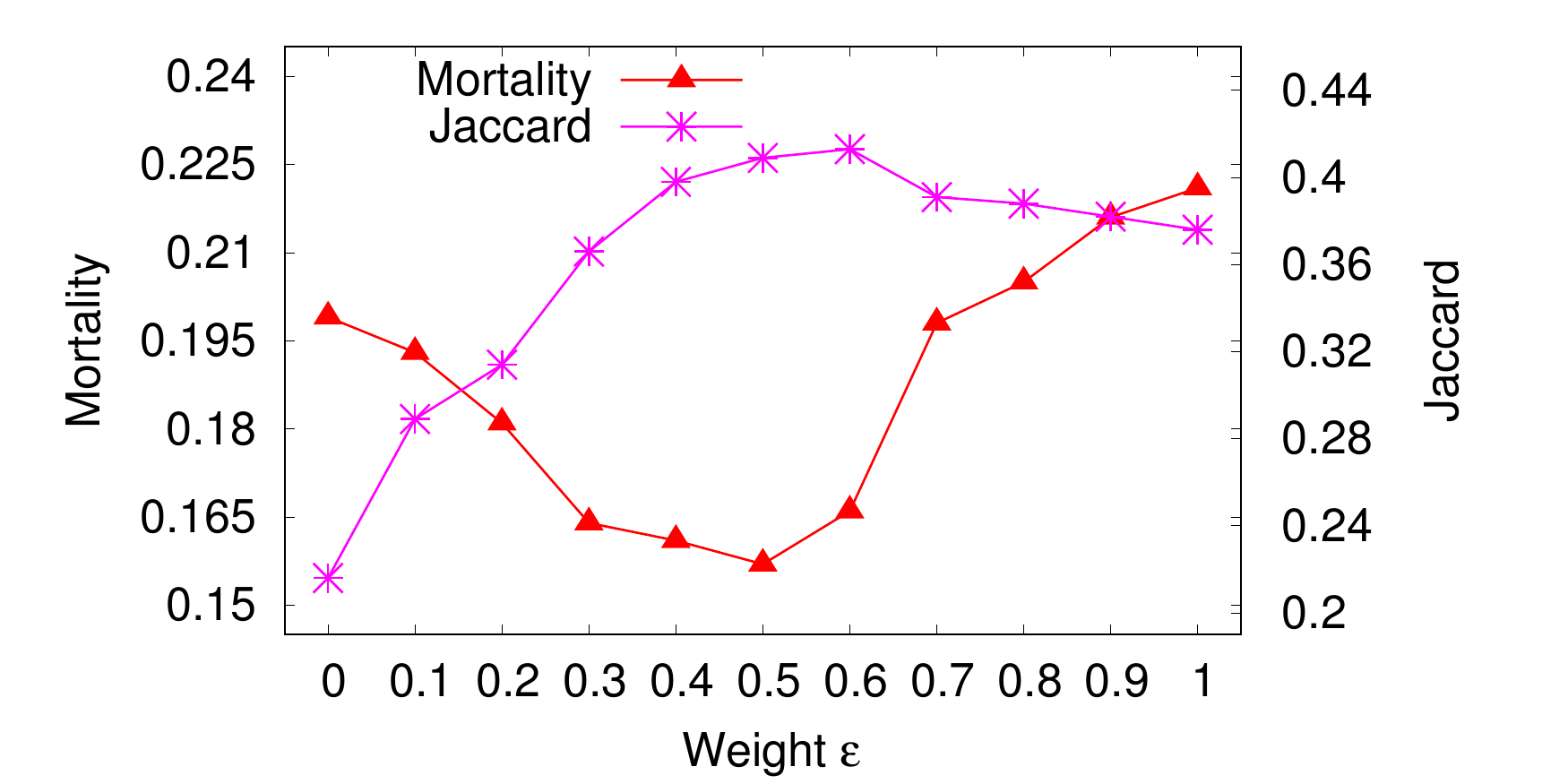}
	%
	%\vspace{-1.5em}
	\caption{The effect of $\epsilon$.}
	\label{fig:epsilon}
	%\vspace{-1.1em}
\end{figure}

\noindent \textbf{Effect of $\epsilon$.}
Figure \ref{fig:epsilon} shows the effectiveness of the weight parameter $\epsilon$ in Equation~\ref{eq:epsilon}, which is used to balance RL and SL.
It achieves the highest Jaccard scores and lowest mortality rates when taking the values of 0.5 and 0.6,
which verifies that SRL-RNN can not only significantly reduce the estimated mortality but also recommend better medications as well.

\section{Conclusion\label{con}}
In this paper, we propose the novel Supervised Reinforcement Learning with Recurrent Neural Network (SRL-RNN) model for DTR,
which combines the indicator signal and evaluation signal through the joint supervised and reinforcement learning.
SRL-RNN incorporates the off-policy actor-critic architecture to discover optimal dynamic treatments
and further adopts RNN to solve the POMDP problem.
The comprehensive experiments on the real world EHR dataset demonstrate SRL-RNN
can reduce the estimated mortality in hospital by up to 4.4\% and provide better medication recommendation as well.

%\begin{table*}
% \caption{Case study of prescription recommendation}
% \label{tab:commands}
%\begin{tabular}{ccccl}
% \toprule
%  \texttt{Dianogsis}&\texttt{days} &\texttt{method} &\texttt{treatment}  \\
% \midrule
%  \multirow{4}{*}{\tabincell{c}{Multiple closed pelvic fractures, \\Infection and inflammatory, \\coagulation defects, Pneumonia,\\ Nonunion,  lower extremities fracture}} &1&LG& \tabincell{c}{Metoprolol, Morphine Sulfate, Lorazepam, Heparin,} \\
% &1& LEAP &0.436\\
% &1&RL&0.495\\
% &1&AC&0.495\\
% \multirow{4}{*}{\tabincell{c}{Perforation of intestine, \\Anticoagulants adverse effects, \\Chronic vascular insufficiency of intestine,\\Vascular disorders, Spleen disease,\\ Defibrination syndrome, Pancreatic disease,\\ Peritoneal abscess, Hematoma complic,\\ thrombosis,Streptococcal septicemia, thrombosis,\\Acute vascular insufficiency of intestine  }} & 2&LG &0.512\\
%&2 &LEAP &0.524\\
%  &2&RL&0.495\\
% &2& AC &0.306\\
%     \bottomrule
% \end{tabular}
%\end{table*}

\begin{table*}
	
	\scriptsize
	\caption{Case study of prescription recommendation.}
	\label{tab:case}
	%\vspace{-1.5em}
	\begin{tabular}{|c|c|c|c|}
		\hline
		\texttt{Diagnosis}&\texttt{day} &\texttt{method} &\texttt{treatment}  \\\hline
		\multirow{8}{*}{\tabincell{c}{Traumatic brain hemorrhage, \\Spleen injury, intrathoracic injury, \\Contusion of lung, \\motor vehicle traffic collision, \\Acute posthemorrhagic anemia}} &  \multirow{4}{*}{1}&Doctor& \tabincell{c}{Potassium Chloride, Magnesium Sulfate, Calcium Gluconate, Phytonadione, Famotidine, Mannitol} \\\cline{3-4}
		&& LG &\tabincell{c}{Potassium Chloride, Acetaminophen,Morphine Sulfate,Docusate Sodium,\\ Bisacodyl, Magnesium Sulfate, Calcium Gluconate, Fentanyl Citrate, Famotidine, Potassium Chloride}\\\cline{3-4}
		&&LEAP&\tabincell{c}{Potassium Chloride, Acetaminophen, Heparin, Magnesium Sulfate,Calcium Gluconate, Fentanyl Citrate, Famotidine}\\\cline{3-4}
		&&SRL-RNN&\tabincell{c}{Potassium Chloride, Acetaminophen, Morphine Sulfate, Magnesium Sulfate, Calcium Gluconate, Fentanyl Citrate, Famotidine}\\\cline{2-4}
		& \multirow{4}{*}{2}&Doctor&\tabincell{c}{Potassium Chloride, Acetaminophen, Magnesium Sulfate, Calcium Gluconate, Fentanyl Citrate, Famotidine}\\\cline{3-4}
		&&LG &\tabincell{c}{Acetaminophen, Morphine Sulfate, Heparin, Docusate Sodium, Bisacodyl, Senna,\\ Oxycodone-Acetaminophen, Ondansetron, Famotidine}\\\cline{3-4}
		&&LEAP &\tabincell{c}{Potassium Chloride, Acetaminophen, Heparin, Magnesium Sulfate, Calcium Gluconate, Fentanyl Citrate, Famotidine}\\\cline{3-4}
		&& SRL-RNN &\tabincell{c}{Potassium Chloride, Acetaminophen, Morphine Sulfate, Magnesium Sulfate, Calcium Gluconate, Fentanyl Citrate, Propofol}
		\\\cline{2-4}\hline
		
		\multirow{8}{*}{\tabincell{c}{Coronary atherosclerosis, pleural effusion,\\ Percutaneous coronary angioplasty, \\Hypopotassemia, pericardium, \\Personal history of malignant neoplasm\\}} &  \multirow{4}{*}{1}&Doctor& \tabincell{c}{Acetaminophen, Morphine Sulfate, Lorazepam, Heparin, Docusate Sodium, Aspirin EC, Zolpidem Tartrate, Amlodipine, Pravastatin} \\\cline{3-4}
		&& LG &\tabincell{c}{Acetaminophen, Metoprolol, Morphine Sulfate, Heparin, Magnesium Sulfate, Senna, \\Aspirin, Pantoprazole, Lisinopril, Atorvastatin, Docusate Sodium, Clopidogrel Bisulfate, Atropine Sulfate, Potassium Chloride}\\\cline{3-4}
		&&LEAP&\tabincell{c}{Acetaminophen, Heparin, Docusate Sodium, Magnesium Sulfate, Potassium Chloride, Acetaminophen}\\\cline{3-4}
		&&SRL-RNN&\tabincell{c}{Acetaminophen, Morphine Sulfate, Lorazepam, Aspirin EC, Docusate Sodium, Pravastatin, Heparin Sodium, Zolpidem Tartrate, }\\\cline{2-4}
		& \multirow{4}{*}{2}&Doctor&\tabincell{c}{Potassium Chloride}\\\cline{3-4}
		&&LG &\tabincell{c}{Acetaminophen, Morphine Sulfate, Heparin, Magnesium Sulfate, Senna, \\Aspirin, Pantoprazole, Lisinopril, Atorvastatin, Docusate Sodium, Clopidogrel Bisulfate, Atropine Sulfate, Potassium Chloride}\\\cline{3-4}
		&&LEAP &\tabincell{c}{Acetaminophen, Heparin, Docusate Sodium, Magnesium Sulfate, Potassium Chloride, Acetaminophen}\\\cline{3-4}
		&& SRL-RNN &\tabincell{c}{Potassium Chloride, Calcium Gluconate, Docusate Sodium}
		\\\cline{2-4}\hline
	\end{tabular}
	%\vspace{-1em}
\end{table*}
\section*{Acknowledgements}

We thank Aniruddh Raghu and Matthieu Komorowski to help us preprocess the data. This work was partially supported by the National Key Research and Development Program of China under Grant No. 2016YFB1000904, NSFC (61702190),
and NSFC-Zhejiang (U1609220).

\bibliographystyle{ACM-Reference-Format}
%\balance
\bibliography{sample-bibliography}

\end{document}